\documentclass[10pt,twocolumn,letterpaper]{article}

\usepackage[pagenumbers]{cvpr} 

\usepackage[utf8]{inputenc} 
\usepackage[T1]{fontenc}    
\usepackage{hyperref}       
\usepackage{url}            
\usepackage{booktabs}       
\usepackage{amsfonts}       
\usepackage{nicefrac}       
\usepackage{microtype}      
\usepackage[dvipsnames]{xcolor}         

\usepackage{wrapfig}        
\usepackage{graphicx}
\usepackage{enumitem}
\usepackage{caption}

\usepackage{amsmath,bm}
\usepackage{dsfont}
\usepackage{amssymb}
\usepackage{amsfonts}
\usepackage{dcolumn} 
\usepackage{tabu}
\usepackage{array}
\usepackage{xspace}
\usepackage{makecell}

\usepackage[bottom]{footmisc}

\usepackage{colortbl}
\usepackage{booktabs, multirow} 
\usepackage{soul}
\usepackage{changepage,threeparttable} 

\PassOptionsToPackage{numbers,compress,sort}{natbib}
\usepackage[numbers]{natbib}

\usepackage{algorithm}
\usepackage{algpseudocode}
\usepackage[vlined,linesnumbered,ruled,algo2e]{algorithm2e}

\usepackage{amsmath}
\usepackage{amssymb}
\usepackage{mathtools}
\usepackage{amsthm}

\usepackage[capitalize,noabbrev]{cleveref}

\theoremstyle{plain}

\theoremstyle{definition}

\theoremstyle{remark}

\usepackage[dvipsnames]{xcolor}

\definecolor{cvprblue}{rgb}{0.21,0.49,0.74}

\def\paperID{373} 
\def\confName{3DV\xspace}
\def\confYear{2026\xspace}

\title{WildSmoke: Ready-to-Use Dynamic 3D Smoke Assets\\from a Single Video in the Wild}

\author{
\begin{tabular}{cccc}
Yuqiu Liu & Jialin Song & Manolis Savva & Wuyang Chen \\
\end{tabular} \\
Simon Fraser University \\
{\tt\small \{yuqiu\_liu, jialin\_song, msavva, wuyang\_chen\}@sfu.ca} \\
}

\begin{document}
\maketitle

\begin{abstract}
We propose a pipeline to extract and reconstruct \textbf{dynamic 3D smoke assets}
from a \textbf{single in-the-wild video}, and further integrate interactive simulation for smoke design and editing.
Recent developments in 3D vision have significantly improved reconstructing and rendering fluid dynamics,
supporting realistic and temporally consistent view synthesis.
However, current fluid reconstructions rely heavily on \ul{carefully controlled clean lab environments}, whereas \ul{real-world videos captured in the wild are largely underexplored}.
We pinpoint three key challenges of reconstructing smoke in real-world videos and design targeted techniques, including smoke extraction with background removal, initialization of smoke particles and camera poses, and inferring multi-view videos.
Our method not only outperforms previous reconstruction and generation methods with high-quality smoke reconstructions (+2.22 average PSNR on wild videos), but also enables diverse and realistic editing of fluid dynamics by simulating our smoke assets.
We provide our models, data, and 4D smoke assets at \href{https://autumnyq.github.io/WildSmoke/}{https://autumnyq.github.io/WildSmoke}.
\end{abstract}

\section{Introduction}

Fluid phenomena,
from the swirling eddies around a high-speed train to the rolling smoke rings emitted by a jet engine,
are ubiquitous in the physical world.
A fundamental yet challenging task is to reconstruct unobserved physical quantities like velocity and density in the full spatiotemporal domain (3+1D), given visual inputs such as 2D photographs or video sequences.
This problem is formally known as inferring \textbf{3D fluid fields} from imagery, i.e., \ul{reconstructing fluid dynamics from visual observations}.
Successfully solving this problem enables applications such as high-fidelity smoke rendering in cinematic special effects~\cite{kim2020lagrangian,wang2024physics} and enhanced diagnostics in industrial fluid systems~\cite{saini2016development,thuerey2020deep,baker2024enginebench}.
Recent breakthroughs in 3D vision have substantially advanced this area.
Key contributions include multi-view benchmarks~\cite{eckert2019scalarflow} that deliver high-quality flow videos with accurately calibrated camera poses.
In addition,
recent methods reconstruct fluid fields from video by jointly optimizing differentiable physics (via physics-based constraints) and neural representations (via rendering losses)~\cite{chu2022physics,guan2022neurofluid,deng2023fluid,deng2023learning,yu2024inferring,gao2025fluidnexus}.

Despite notable advances, most existing approaches still rely heavily on \textbf{controlled multi-view} recordings for reconstructing fluid phenomena, which \textbf{differ substantially from real-world smoke in the wild}.
Datasets such as ScalarFlow~\cite{eckert2019scalarflow}, TomoFluid~\cite{zang2020tomofluid}, and FluidNexus~\cite{gao2025fluidnexus} rely on fixed, precisely calibrated cameras, controlled smoke generation in laboratory settings, and pre-captured simplistic backgrounds.
Such conditions are impractical for in-the-wild smoke, which is often recorded with a single moving camera (e.g., handheld or drone footage), 
with cluttered backgrounds and unpredictable camera motion.
At the same time, demand for ready-to-use 3D assets has surged across graphics and simulation applications~\cite{poole2022dreamfusion,xiang2024structured,zhao2025hunyuan3d}.
\textbf{Dynamic 3D smoke assets are particularly valuable} for downstream visual effects editing and real-time simulations~\cite{bai2020dynamic,chu2021learning,zhao2025vid2fluid}, owing to their intricate motion and evolving volumetric properties.
Yet, despite their importance, the generation of dynamic 3D smoke assets remains largely underexplored.

Motivated by the above challenges, we ask:

\begin{center}
\fbox{
    \parbox{0.9\linewidth}{
        \textit{Given a single in-the-wild video, how can we recover the underlying fluid field and enable downstream applications using the reconstructed smoke assets?}

    }
}
\end{center}

In this work, we aim to reconstruct ready-to-use dynamic 3D smoke assets from a single in-the-wild video.
We identify three key challenges of reconstructing smoke fields from the wild (Figure~\ref{fig:challenges}, Section~\ref{sec:challenges}): noisy backgrounds, unknown camera poses, and coupled camera viewpoints and timesteps.
Correspondingly, we introduce three key techniques:
1) In-context smoke segmentation, plus dehazing for light smoke background removal;
2) Initializations of smoke particles and pose estimation;
3) Decoupling spatial and temporal camera trajectories by multi-view generation and local pose perturbation. 
More importantly,
we demonstrate practical applications of our reconstructed smoke assets through fluid simulation.
Our main contributions are summarized as follows:

\begin{enumerate}[leftmargin=*]
    \item
    We design a comprehensive pipeline that extracts and reconstructs dynamic 3D smoke assets from diverse and noisy real-world videos (Section~\ref{sec:methods}).

    \item
    Compared with current fluid field reconstruction or 3D generation methods, our pipeline can faithfully reconstruct accurate and dynamic 3D smoke assets from a single in-the-wild video, achieving an average improvement of +2.22 dB PSNR on in-the-wild videos
    (Section~\ref{sec:exp_real}).

    \item
    Our smoke assets are ready-to-use: we support realistic editing of smoke through interactive fluid simulations (Section~\ref{sec:method_simulation} and ~\ref{sec:exp_simulation}).
\end{enumerate}

\section{Background}

\subsection{Smoke Reconstruction from Videos}

Given a single-view video capturing upward-rising smoke, our goal is to reconstruct a dynamic 3D representation of the smoke field over time ($t = 1,\cdots,T$) that is coherent, view-consistent, and editable. 
Following FluidNexus~\cite{gao2025fluidnexus},
we represent fluid with two types of particles.
\begin{enumerate}[leftmargin=*]
    \item
3D physical particles for positions and velocities: $\mathbf{p}^\text{phy}_t, \mathbf{u}_t \in \mathbb{R}^{N_t^\text{phy}\times 3}$, where $N_t^\text{phy}$ is the number of physical particles at $t$.
The density field $\rho_t: \mathbb{R}^3\mapsto\mathbb{R}$ and velocity field $\mathbf{V}_t: \mathbb{R}^3\mapsto\mathbb{R}^3$
can be further mapped from particles to grids via kernel-weighted interpolation~\cite{hu2019taichi,hu2019difftaichi,hu2021quantaichi,muller2003particle,gao2025fluidnexus}.

    \item
$N_t^\text{vis}$ visual particles (grayscale) at $t$ with attributes:
$\{\mathbf{p}^\text{vis}_{t}\in\mathbb{R}^{N^\text{vis}_t\times 3}, \mathbf{c}_{t}\in\mathbb{R}^{N^\text{vis}_t}, \mathbf{s}_t\in\mathbb{R}^{N^\text{vis}_t\times 3}, \mathbf{o}_t\in\mathbb{R}^{N^\text{vis}_t}, \mathbf{r}_t\in\mathbb{R}^{N^\text{vis}_t\times 4}\}$, 
representing position, color, scale, opacity, and rotation, respectively.
\end{enumerate}

\subsection{Smoke Reconstruction in the Wild: Challenges}
\label{sec:challenges}

\begin{figure}
	\centering
	\includegraphics[width=0.46\textwidth]{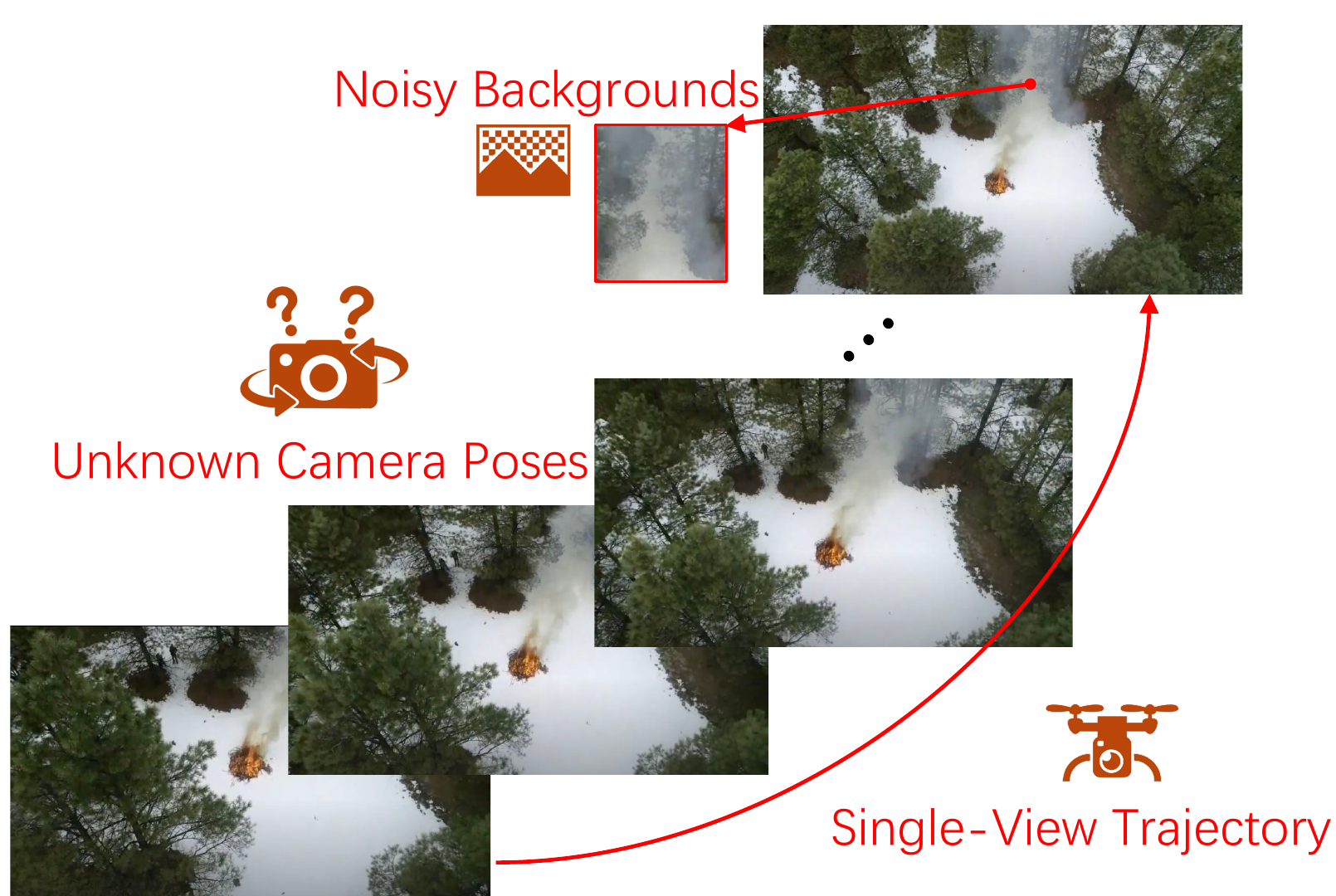}
    \vspace{-0.5em}
    \captionsetup{font=small}
	\caption{Challenges of smoke reconstruction from a single in-the-wild video:
    1) noisy backgrounds and boundaries;
    2) unknown camera poses;
    3) single video with coupled camera viewpoints and timesteps.
}
    \label{fig:challenges}
\vspace{-1em}
\end{figure}

We identify three challenges when reconstructing smoke from a single video in the wild (Figure~\ref{fig:challenges}):

\vspace{-1.2 em}
\paragraph{Noisy Backgrounds and Boundaries.} High-quality videos of real smoke like ScalarFlow~\cite{eckert2019scalarflow}, TomoFluid~\cite{zang2020tomofluid}, and FluidNexus~\cite{gao2025fluidnexus} were collected with carefully controlled backgrounds and further post-processed to remove environmental noise.
However, videos in the wild are inevitably contaminated by complex backgrounds visible through semi-transparent smoke.

\vspace{-1.2 em}
\paragraph{Unknown Camera Poses.}
Although videos of smoke in the wild are captured with moving cameras,
their camera poses are typically neither measured nor released with the footage.

\vspace{-1.2 em}
\paragraph{Single-Camera Trajectory.}
As a video of smoke in the wild typically
provides only one camera trajectory over time,
spatial viewpoints and temporal frames become inherently entangled: there is an (approximately) one-to-one correspondence between camera viewpoints and timesteps.

\section{Methods}
\label{sec:methods}

\begin{figure*}[h!]
	\centering
	\includegraphics[width=0.9\textwidth]{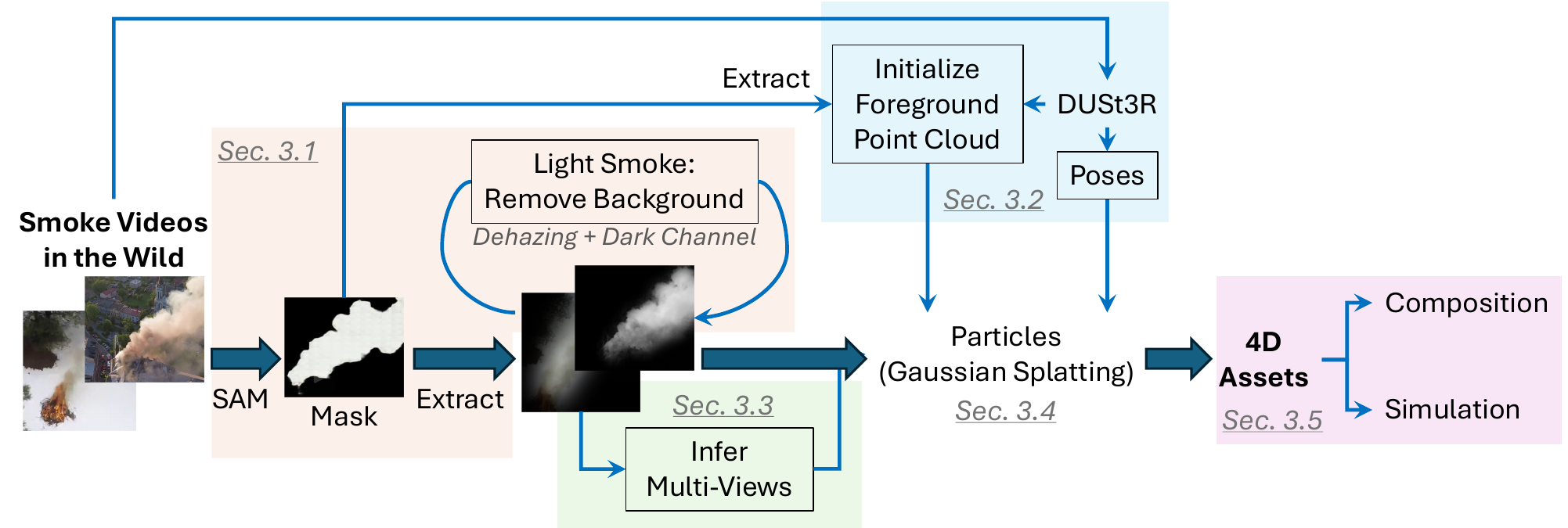}
    \vspace{-0.5em}
    \captionsetup{font=small}
	\caption{To reconstruct smoke from a single video in the wild, our pipeline includes five steps:
    1) Smoke extraction, with background removed for light smoke due to translucence (Sec.~\ref{sec:method_smoke_extraction});
    2) Pose estimation and coarse initialization for 3D point cloud (Sec.~\ref{sec:method_pose_estimation});
    3) Inferring multi-views for smoke (Sec.~\ref{sec:method_multi_view});
    4) Training smoke particles (Sec.~\ref{sec:method_training});
    5) composition and simulation of 4D smoke assets (Sec.~\ref{sec:method_simulation}).
    ``SAM'': segment anything~\cite{kirillov2023segment}. ``DUSt3R''~\cite{wang2024dust3r}.}
    \label{fig:overview}
\end{figure*}

To address the above challenges, we propose a unified pipeline to reconstruct clean, background-free smoke from unconstrained real-world videos, as shown in Figure~\ref{fig:overview}.

\subsection{Smoke Extraction}
\label{sec:method_smoke_extraction}

In real-world scenarios, smoke often exhibits complex, irregular boundaries and is intertwined with  significant noise from backgrounds.
To address this issue, we propose a pipeline that automatically extracts clean smoke regions from a single in-the-wild video.

\vspace{-1 em}
\paragraph{(a) Smoke Mask Extraction.}
We first segment the smoke and obtain a binary mask sequence $\mathbf{M}=\{M_t\}$ over all timesteps $t$ in the video. 
Because off-the-shelf semantic segmentation does not generalize well to in-the-wild smoke, we adopt one-shot learning,
using SAM~\cite{kirillov2023segment} for annotation and SegGPT~\cite{wang2023images} for propagation.
For each video, we select an early reference frame $I_{\mathrm{ref}}$, interactively annotate its smoke with SAM to obtain $M_{\mathrm{ref}}$\footnote{We use an online SAM-based annotator \href{https://roboflow.com/}{https://roboflow.com/}.}, and then feed the pair $(I_{\mathrm{ref}}, M_{\mathrm{ref}})$ to SegGPT~\cite{wang2023images} to perform one-shot inference and produce masks $M_t$ for the remaining frames $I_t$. 
This combines SAM’s high-quality per-frame annotation with SegGPT’s frame-wise generalization.

The boundary of segmented smoke could still be noisy.
We apply a Gaussian filter $f_g$ to smooth the boundary of inferred masks, yielding a more natural smoke contour.
The smoke areas are extracted via $\tilde{S}_t = f_g(M_t) \cdot I_t$.
As shown in Figure~\ref{fig:background_removal}(a),
this in-context segmentation can reliably segment smoke boundaries. 
See Appendix~\ref{sec:smoke_sample_supp} in the supplement for more smoke segmentation examples.

At this stage, it is necessary to further distinguish dense from light smoke. We categorize them based on the visibility of the background.
For dense smoke, the background is completely occluded due to high optical density (low transmittance). In this case, background contamination is negligible, and the smoke can be directly treated as the masked region.
However, for the light smoke, the background remains partially visible (nonzero transmittance). Here, a dehazing step is required to remove background leakage before smoke reconstruction.

\vspace{-1 em}
\paragraph{(b) Dehazing for Background Removal.}
Similar to fog or haze, background objects remain visible through light smoke due to light transmission, introducing artifacts during reconstruction.
For light smoke, we fine-tune a pretrained DehazeFormer~\cite{song2023vision} to isolate the smoke and remove the background.
We construct a fine-tuning dataset of synthetic smoky inputs $\tilde{I}$ (pixels normalized to $[0, 1]$) by blending coarsely extracted smoke $\tilde{S}$ with video frames of clean background without smoke $I^\text{clean}$, which serve as supervision targets:
\begin{equation}
\tilde{I} = I^\text{clean} \cdot \tilde{T} + A \cdot \tilde{S}, 
\label{eq:haze}
\end{equation}
where $\tilde{T} = 1-\tilde{S}$ is the coarse transmission map, representing the amount of background light that reaches the camera through the smoke.
We estimate the atmospheric light $A$ using the dark channel prior~\cite{he2010single}.
As shown in Figure~\ref{fig:background_removal} (c), the fine-tuned DehazeFormer can
remove the foreground smoke and recover the background $\tilde{I}^\text{clean}$.

For dense smoke where background contamination is negligible, the extracted smoke $\tilde{S}$ from masks can be used directly without this dehazing step.

\vspace{-1 em}
\paragraph{(c) Extract Clean Smoke.}
For light smoke, we extract the clean foreground smoke based on Equation (\ref{eq:haze}):
\begin{equation}
    S = 1-\frac{I - A}{\tilde{I}^\text{clean} - A},
\end{equation}
where $I$ is the smoky frame in the original video, and $\tilde{I}^\text{clean}$ is the background recovered by DehazeFormer.
The resulting foreground-smoke images are then used for smoke-field reconstruction.

\begin{figure*}[h!]
\vspace{-0.5em}
	\centering
	\includegraphics[width=0.85\textwidth]{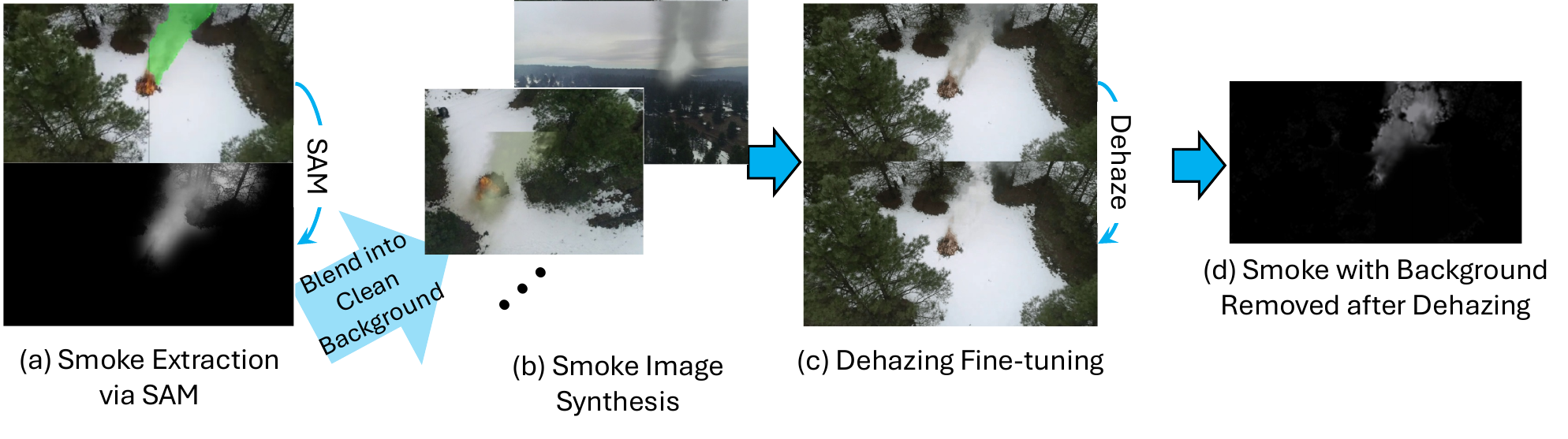}
    \vspace{-0.5em}
    \captionsetup{font=small}
	\caption{Smoke extraction, with background removal for light smoke.}
    \label{fig:background_removal}
    \vspace{-1em}
\end{figure*}

\subsection{Pose Estimation and Coarse Geometry}
\label{sec:method_pose_estimation}

To obtain camera intrinsics and extrinsics for frames, we initialize camera poses and focal lengths with estimates from the pretrained DUSt3R~\cite{wang2024dust3r}.
We further initialize both physical and visual particles from a sparse 3D point cloud $\{\mathbf{p}_{i}\}_{i=1}^{N}$ estimated by DUSt3R, which provides a rough geometry of the scene, including both smoke and background regions. 
We filter the DUSt3R-generated 3D point cloud using the extracted smoke masks, retaining only the smoke-relevant points to initialize our particles.
Although DUSt3R produces coarse results, this initialization provides meaningful spatial priors for optimizing our particles; without it, training may fail to converge.
See more details in Appendix~\ref{sec:pose_estimation_supp} in the supplement.

\subsection{Inferring Multi-View Videos}
\label{sec:method_multi_view}

In wild smoke videos,
cameras typically follow a single trajectory.
For example, a drone may fly from the ground upwards along with the smoke plume to capture the video.
Although the camera may span a wide spatiotemporal extent, camera viewpoints and timesteps are highly coupled, i.e., each camera viewpoint is associated with a unique timestep, and vice versa.
This coupling can
cause overfitting and degrade novel-view synthesis for unseen viewpoints or timesteps.

To decouple the spatiotemporal camera trajectory,
in our work we employ generative multi-view synthesis.
To obtain multi-view supervision from single-view input, we use pretrained SV4D 2.0~\cite{yao2025sv4d} to generate smoke videos from novel views.
We choose azimuth angles of $[-10^\circ, 10^\circ, 20^\circ, 30^\circ]$
relative to the pose of the current frame
as our novel viewpoints.
Since SV4D 2.0 does not support long temporally consistent video generation, we split our video sequences into short clips, and overlap one frame between neighboring clips, ensuring valid conditioning for every novel-view segment.
Per-frame camera-to-world poses are obtained by applying the angle set to the DUSt3R-initialized poses.

Moreover, to address the unreliable frames produced by SV4D at later timesteps,
we apply an exponentially decaying weight over the frame index.
This strategy progressively down-weights the influence of later frames generated by SV4D 2.0 during Gaussian-particle training.
See Appendix~\ref{sec:generated_vid_supp} and~\ref{sec:weight_decay_supp} for more details.

\subsection{Training Gaussian Particles}
\label{sec:method_training}

Following the training strategy of FluidNexus~\cite{gao2025fluidnexus}, we separately train visual and physical particle representations.
Particles are optimized by minimizing photometric errors between input frames and rendered views using 3D Gaussian Splatting~\citep{kerbl20233d}.
Physical particles are further regularized using position-based fluid (PBF) simulation~\cite{macklin2013position,macklin2014unified} with an incompressibility constraint. 

\vspace{-1.2 em}
\paragraph{Local Pose Perturbation.}
To further disentangle spatial and temporal viewpoints in single-camera trajectories from in-the-wild videos, we not only train with our generated multi-view trajectories (Sec.~\ref{sec:method_multi_view}), but also progressively enrich the view trajectory by perturbing local camera poses.
Specifically, after the particles at time $t$ converge, we introduce perturbed viewpoints by shifting the camera pose forward by $\Delta t$ along the trajectory (modulo the sequence length $T$).
Concretely, together with the original pose $(\mathbf{R}_t, t)$, we also include $(\mathbf{R}_{(t+\Delta t)~\text{mod}~T}, t)$ as an input and its corresponding rendering result as the target.
Here, $\mathbf{R}$ is the rotation matrix of the camera's extrinsics; $\Delta t \ll T$, i.e., we perturb within a very short period relative to the whole temporal domain.
Essentially,
we effectively decouple viewpoint and timestep by associating pose $\mathbf{R}_{(t+\Delta t)~\text{mod}~T}$ with timestep $t$
via ``local pose perturbation.''
Since $\Delta t$ is small, perturbed novel viewpoints still reside in neighborhoods of the original video trajectory.
In practice, we progressively increase $\Delta t$ from 2 to 4 during training.
We refer readers to Appendix~\ref{sec:tech_supp} in the supplement for more training details.

\begin{figure}
\vspace{-1em}
	\centering
	\includegraphics[width=0.45\textwidth]{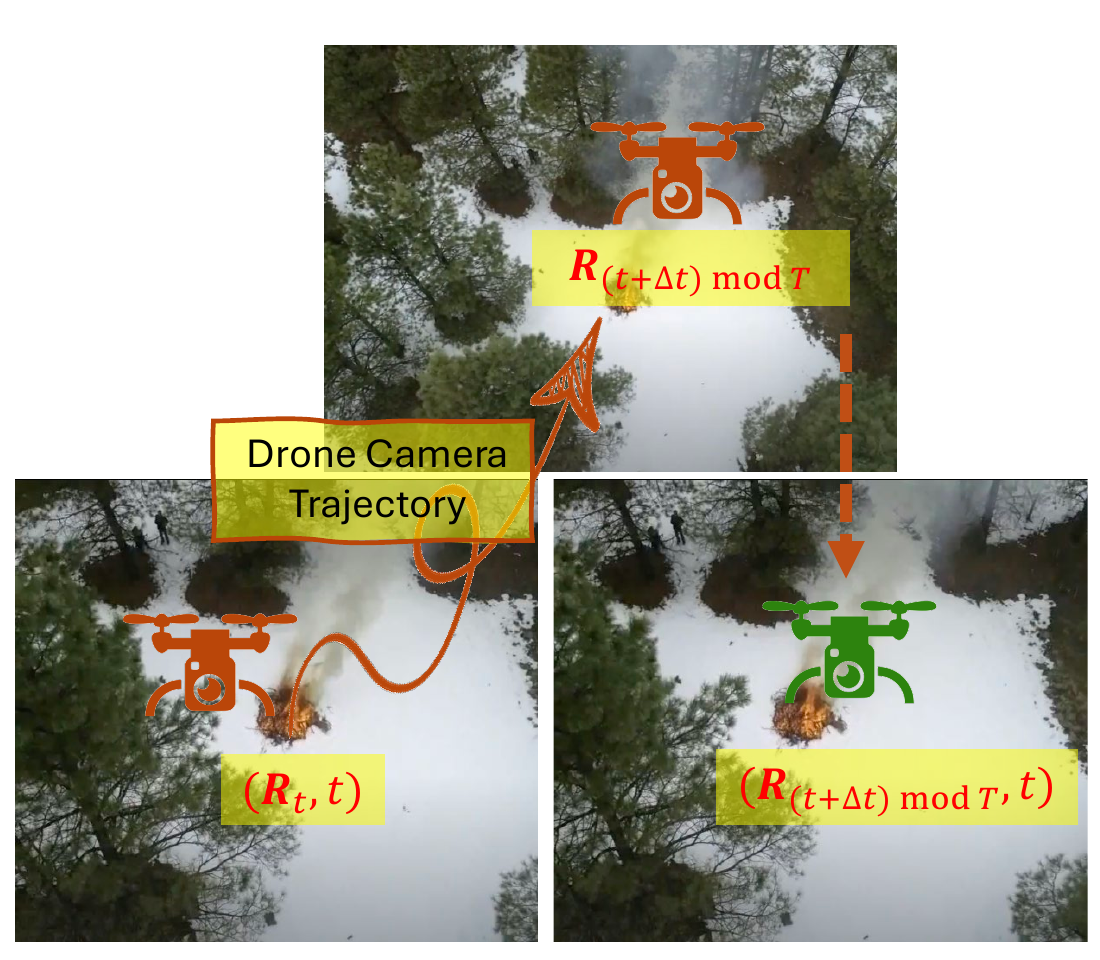}
    \vspace{-0.5em}
    \captionsetup{font=small}
	\caption{Local pose perturbation.
    Given the original camera pose $(\mathbf{R}_t, t)$ on a single-video trajectory, we additionally introduce a perturbed viewpoint by shifting the camera pose by $\Delta t$ along the trajectory with wrap-around $(t+\Delta t)~\text{mod}~T$.
    For this perturbed viewpoint, the rendered image serves as supervision. Both original and perturbed samples are used during following training.
    }
    \label{fig:spatiotemporal_decomposition}
    \vspace{-1em}
\end{figure}

\subsection{Ready-to-Use 4D Smoke Assets}
\label{sec:method_simulation}

A key use case of our reconstructed smoke assets is visual-effects (VFX) editing of novel smoke scenes.
In our work, we consider simulating smoke interactions using PhiFlow~\cite{holl2024phiflow}.
The simulation is initialized with the reconstructed density field $\rho$ from visual particles, and velocity field $\mathbf{V}$ from physical particles:
\vspace{-1.5em}
\paragraph{Density Field from Visual Particles.} For each visual particle $i$ with center $\mathbf{p}^\text{vis}_{t,i}$, scale $\mathbf{s}_{t,i}$, rotation $\mathbf{r}_{t,i}$, and opacity $\mathbf{o}_{t,i}$,  
we define a Gaussian kernel  
\begin{equation}
\phi_{t,i}(\mathbf{x}) = \exp\!\left(-\frac{1}{2} (\mathbf{x} - \mathbf{p}^\text{vis}_{t,i})^\top \boldsymbol{\Sigma}_{t,i}^{-1} (\mathbf{x} - \mathbf{p}^\text{vis}_{t,i}) \right),
\end{equation}
where $\boldsymbol{\Sigma}_{t,i} = R(\mathbf{r}_{t,i}) \, \mathrm{diag}(\mathbf{s}_{t,i}^2) \, R(\mathbf{r}_{t,i})^\top$, $\mathbf{r}_{t,i}\in\mathbb{R}^4$ is the unit quaternion representing the particle rotation, 
and $R(\mathbf{r}_{t,i})\in\mathbb{R}^{3\times 3}$ is the corresponding rotation matrix obtained from $\mathbf{r}_{t,i}$.
The density field is the weighted sum of all particle kernels:
\begin{equation}
\rho_{t}(\mathbf{x}) = \sum_{i=1}^{N^\text{vis}_t} \mathbf{o}_{t,i} \, \phi_{t,i}(\mathbf{x}) .
\end{equation}

\vspace{-1.2em}
\paragraph{Velocity Field from Physical Particles.}  
Given a physical particle with position $\mathbf{p}^\text{phy}_{t,i}$ and velocity $\mathbf{u}_{t,i}$,  
we map them to the grid. The Gaussian kernel $\phi^\prime_{t,i}$ for the velocity field has the same form as $\phi_{t,i}$ but is centered at $\mathbf{p}^\text{phy}_{t,i}$:
\begin{equation}
\mathbf{V}_t(\mathbf{x}) = \frac{\sum_{i=1}^{N^\text{phy}_t} \phi^\prime_{t,i}(\mathbf{x}) \, \mathbf{u}_{t,i}}{\sum_{i=1}^{N^\text{phy}_t} \phi^\prime_{t,i}(\mathbf{x}) + \varepsilon},
\end{equation}
here we use $\varepsilon=10^{-8}$. 
To ensure spatial consistency between the density and velocity fields,  
both visual and physical splatting are restricted to the same bounding region defined by the visual particles.

After the initialization of density and velocity fields, we continue the fluid dynamics in PhiFlow~\cite{holl2024phiflow} with the standard MacCormack semi-Lagrangian method~\cite{selle2008unconditionally} for advection and the projection-based pressure-Poisson solver for incompressible flow.
We consider either external wind forces or an inserted obstacle to interact with the smoke.

\section{Experiments}
\label{sec:exp}

\subsection{Settings}
\label{sec:settings}

\paragraph{Datasets.}

\begin{figure*}[t!]
\centering
\vspace{-1em}
\includegraphics[width=1\linewidth]{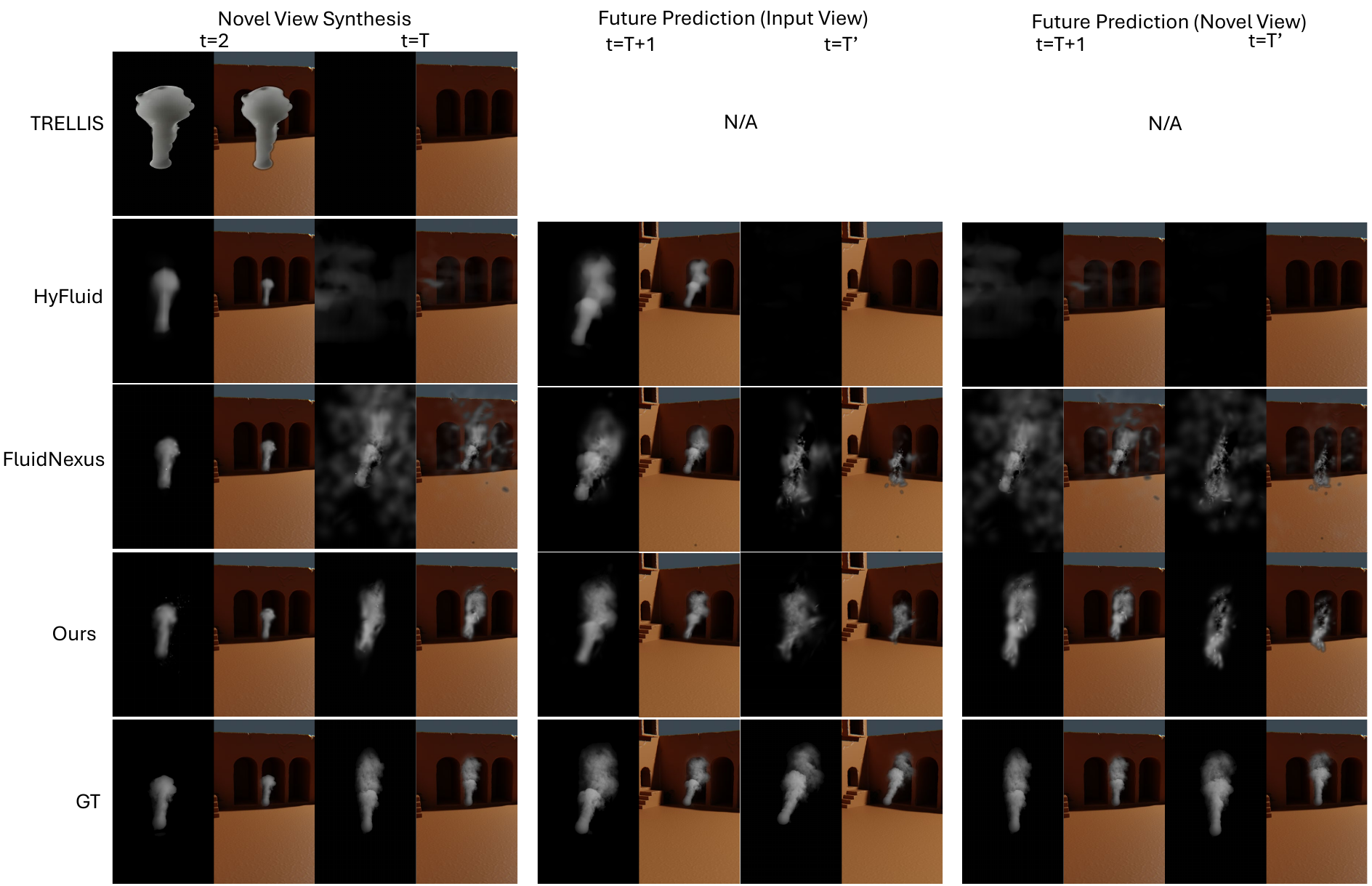}
\centering 
\vspace{-1em}
\captionsetup{font=small}
\caption{Visualization of novel view synthesis and future predictions on \ul{synthetic smoke videos}.
\textit{Novel view} uses the camera pose at $t=1$, and \textit{input view} means the camera poses along the source video.
Training uses $T=240$ frames, and the unseen future extends to $T'=270$ frames. \textit{GT}: Ground Truth.
} 
\vspace{-0.5em}
\label{fig:synthetic_visualization}
\end{figure*}

We evaluate our pipeline on both synthetic and real-world collections of smoke, assessing the reconstruction quality using view rendering.

\begin{itemize}[leftmargin=*]
    \item We \textbf{synthesize} two high-resolution and photorealistic smoke videos by combining a 4D smoke VDB sequence with a real 3D scene. These paired videos, rendered with distinct camera trajectories, allow direct comparison between our reconstructed results and the ground-truth.
    Specifically, we download a synthetic 4D smoke VDB sequence and a 3D scene from \href{https://www.cgtrader.com/3d-models}{CGTrader}\footnote{\href{https://www.cgtrader.com/3d-models}{https://www.cgtrader.com/3d-models}}, combine them and render two videos in Blender\footnote{\href{https://www.blender.org/}{https://www.blender.org}} with two camera trajectories.
    One trajectory is used for training and another for evaluation. See Appendix~\ref{sec:syn_data_supp} for more rendered samples.

    \item We also evaluate on two \textbf{real-world} testbeds: 1) FLAME dataset~\cite{shamsoshoara2021aerial} is a fire-imaging dataset collected by drones during a prescribed burning of piled detritus in an Arizona pine forest. The dataset includes video recordings and thermal heatmaps captured by infrared cameras. 2) We further collect three videos of smoke from \href{https://pixabay.com/}{Pixabay}\footnote{Pixabay shares royalty-free images, videos, audio, and other media. All content is released by Pixabay under the Content License, which makes it safe to use without asking for permission or giving credit to the artist. See their license at \href{https://pixabay.com/service/terms/}{https://pixabay.com/service/terms}}, covering diverse scenarios.

\end{itemize}

To unify our training and evaluation settings, we standardize all our videos to 270 frames; training timesteps are \(t=1,\dots,T\) with \(T=240\), and unseen future timesteps are \(t=T+1,\dots,T'\) with $T'=270$.

We measure view changes along the video trajectory via the camera's relative rotation angle
$\Delta \theta_{t_1,t_2}=\arccos \left(\frac{\operatorname{tr}\left(\Delta \mathbf{R}_{t_1,t_2}\right)-1}{2}\right)$,
where
$\Delta \mathbf{R}_{t_1,t_2} = \mathbf{R}_{t_2}^\top \mathbf{R}_{t_1}$ ($\mathbf{R}$ is the rotation matrix of the camera's extrinsics).
In our synthetic data setting, the camera undergoes a rotation of \(53^\circ\) from \(t=1\) to \(t=T\), which is larger than the \(7^\circ\) rotation during the future steps from \(t=T+1\) to \(t=T'\).
Smoke in our videos is always centered in the scene.

\vspace{-1.em}
\paragraph{Tasks.}

As explained by $\Delta \theta_{1,T}$ and $\Delta \theta_{T+1,T'}$ above,
larger pose changes occur within the training timesteps than in the future steps.
Synthesizing frames from the fixed viewpoint at the beginning of video (\(t=1\)) is more challenging than following the camera pose trajectory.
Therefore, in our evaluation, the \textbf{novel view} is defined with camera pose at $t=1$, and the \textbf{input view} follows the ground-truth camera trajectory.
Following~\cite{yu2024inferring}, we consider the following two tasks:

\begin{itemize}[leftmargin=*]

\item
\textbf{Novel view synthesis:}
We render smoke views from the fixed novel viewpoint over the training timesteps.
Specifically, we fix the camera pose at
$t=1$ and synthesize novel views through $t=2,\cdots,T$.
We study novel view synthesis only on our synthetic dataset,
due to the lack of ground-truth multi-view videos on real-world videos (FLAME and Pixabay).

\item
\textbf{Future prediction:}
We extrapolate the fluid dynamics into the future steps $t=T+1,\cdots,T'$.
No model is ever trained with ground-truth future frames from videos.
On synthetic videos, we study both quantitative and visual results, and predict futures for both input view (i.e. follow the ground-truth camera pose trajectory during $t=T+1,\cdots,T'$) and novel view (fixed camera pose at $t=1$).
Due to the lack of ground-truth multi-view videos, on real-world videos (FLAME and Pixabay), we can only study future predictions based on the input view.

\end{itemize}

During inference, the learned velocity field is used to advect (evolve) the visual particles for future prediction.
We refer the reader to~\cite{yu2024inferring} for more details about these tasks.

\vspace{-1.em}
\paragraph{Evaluation Metrics.}
We report the peak signal-to-noise ratio (PSNR) averaged over frames, and defer the structural similarity index measure (SSIM) and the perceptual metric LPIPS~\cite{zhang2018unreasonable} in Appendix~\ref{sec:ssim_lpips_supp}. These metrics are also widely adopted in prior deblurring works~\cite{nah2017deep,kupyn2018deblurgan}.

\vspace{-1.em}
\paragraph{Baselines.}
We compare with both reconstruction and generation methods, including
HyFluid~\cite{yu2024inferring},
FluidNexus~\cite{gao2025fluidnexus}, and
Trellis~\cite{xiang2024structured}.
We follow FluidNexus to train with the default grayscale input setting.

\begin{table}[h!]
\centering
\captionsetup{font=small}

\caption{Comparing PSNR (higher is better) of smoke reconstruction by different methods on the \ul{synthetic dataset}.
\textit{Novel view} uses the camera pose at $t=1$, and \textit{input view} means the camera poses along the source video.}
\resizebox{0.48\textwidth}{!}{
\addtolength{\tabcolsep}{-0.35em}
\begin{tabular}{lccccccccc}
\toprule
Methods & \begin{tabular}{@{}c@{}}Novel View\\Synthesis\end{tabular} & \begin{tabular}{@{}c@{}}Future Prediction\\(Input View)\end{tabular} & \begin{tabular}{@{}c@{}}Future Prediction\\(Novel View)\end{tabular} \\ \midrule 
Trellis~\cite{xiang2024structured} & 19.98  & - & - \\  
HyFluid~\cite{yu2024inferring} & 24.26  & 22.54 & 22.18 \\  
FluidNexus~\cite{gao2025fluidnexus} & 29.26  & 23.61 & 21.54 \\  
Ours            & \textbf{29.78} & \textbf{25.26} & \textbf{25.04} \\ \bottomrule
\end{tabular}
}
\vspace{-.5em}
\label{table:psnr_synthetic}
\end{table}

\begin{table}[h!]
\centering
\captionsetup{font=small}
\caption{Cumulative ablation study (PSNR) on the \ul{synthetic smoke video}. \textit{Novel view} uses the camera pose at $t=1$, and \textit{input view} means the camera poses along the source video.
}

\resizebox{0.47\textwidth}{!}{
\addtolength{\tabcolsep}{-0.35em}
\begin{tabular}{lccc}
\toprule
 & \begin{tabular}{@{}c@{}}Novel View\\Synthesis\end{tabular} & \begin{tabular}{@{}c@{}}Future Prediction\\(Input View)\end{tabular} & \begin{tabular}{@{}c@{}}Future Prediction\\(Novel View)\end{tabular} \\ \midrule
Baseline & 22.55 & 16.69 & 18.17 \\
+ Smoke Extraction & 29.26 & 23.61 & 21.54 \\
+ DUSt3R Init. & 29.48 & 26.45 & 22.92 \\
+ Local Perturbation & 29.77 & \textbf{26.85} & 23.59 \\
+ Multi-Views (Ours) & \textbf{29.78} & 25.26 & \textbf{25.04} \\ \bottomrule
\end{tabular}
}
\label{table:ablation_synthetic}
\end{table}

\subsection{Synthetic Data}
\label{sec:exp_synthetic}

\paragraph{Results.}
We first show quantitative results in Table~\ref{table:psnr_synthetic} and visualizations in Figure~\ref{fig:synthetic_visualization}.
Trellis~\cite{xiang2024structured} fits a static 3D asset per frame (image-to-3D), estimating neither time-varying particles nor velocities; hence future prediction is ill-defined. Moreover, per-frame inconsistency in the novel-view synthesis setting causes the smoke to progressively fade, yielding a black frame at time $T$.
Compared with HyFluid~\cite{yu2024inferring} and FluidNexus~\cite{gao2025fluidnexus}, our pipeline achieves higher PSNR and more stable visualizations.

\vspace{-1.2em}
\paragraph{Ablation Study.}
We further provide ablation studies in Table~\ref{table:ablation_synthetic}; see visualization results in Appendix~\ref{sec:ablation_supp}.
Smoke extraction (segmentation), initializing poses/particles, and adding inferred multi-view supervision progressively improve PSNR and visual quality.
Note that, due to significant spatiotemporal deviations, the task of \textit{future prediction at novel view} is substantially more challenging than either \textit{novel view synthesis} or \textit{future prediction at input view}.
As a result, \textit{future prediction at novel view} demands additional geometric cues, even coarse ones generated by SV4D, and therefore benefits most (+1.45 PSNR from 23.59 to 25.04) from incorporating ``+ Multi-Views.''

\begin{figure}[h]
\centering
\includegraphics[width=1.0\linewidth]{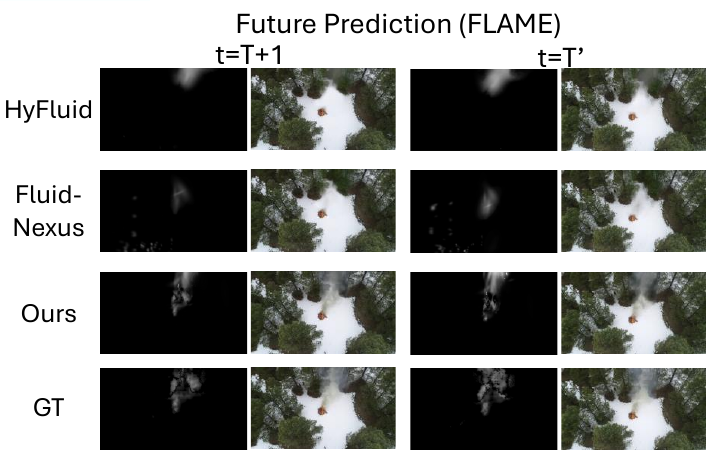}
\centering 
\vspace{-0.5em}
\captionsetup{font=small}
\caption{Visualization of future predictions (input view) on the \ul{FLAME dataset}~\cite{shamsoshoara2021aerial}.
Training uses $T=240$ frames, and the unseen future extends to $T'=270$ frames. \textit{GT}: Ground Truth.
} 
\vspace{-0.5em}
\label{fig:flame_visualization}
\end{figure}

\begin{figure}[t!]
\centering
\includegraphics[width=1.0\linewidth]{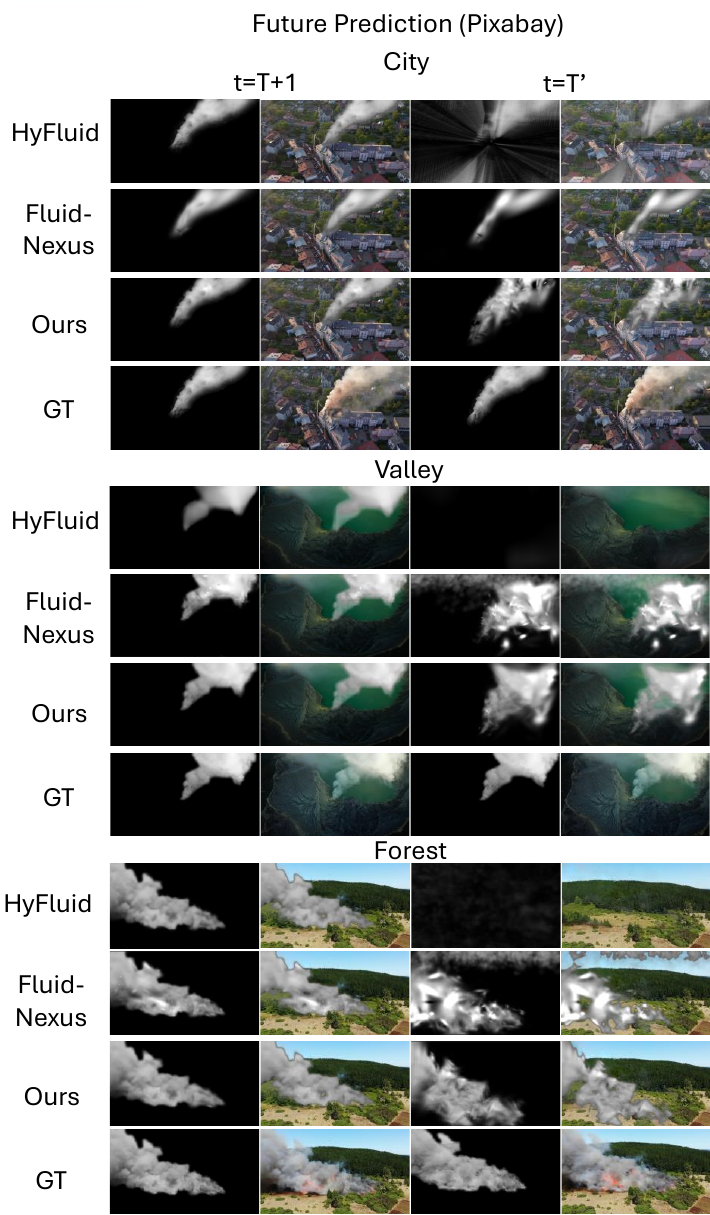}
\centering 
\captionsetup{font=small}
\caption{Visualization of future predictions (input view) on videos collected from \href{https://pixabay.com/}{\ul{Pixabay}}.
Training uses $T=240$ frames, and the unseen future extends to $T'=270$ frames. \textit{GT}: Ground Truth.
} 
\vspace{-0.5em}
\label{fig:pixabay_visualization}
\end{figure}

\begin{table}[h!]
\centering
\captionsetup{font=small}
\caption{Comparing PSNR (higher is better) of smoke reconstruction by different methods on the \ul{FLAME dataset}~\cite{shamsoshoara2021aerial}.
}

\vspace{-0.5em}
\resizebox{0.36\textwidth}{!}{
\addtolength{\tabcolsep}{-0.35em}
\begin{tabular}{lc}
\toprule
 & Future Prediction (Input View) \\ \midrule
HyFluid~\cite{yu2024inferring} & 21.67 \\
FluidNexus~\cite{gao2025fluidnexus} & 21.78 \\
Ours & \textbf{22.88} \\ \bottomrule
\end{tabular}
}

\vspace{-0.5 em}
\label{table:flame}
\end{table}

\begin{table}[h!]
\centering
\captionsetup{font=small}
\caption{Comparing PSNR (higher is better) of smoke reconstruction by different methods on videos collected from \href{https://pixabay.com/}{\ul{Pixabay}}.}

\vspace{-0.5em}
\resizebox{0.38\textwidth}{!}{
\begin{tabular}{lccc}
\toprule
 &\multicolumn{3}{c}{Future Prediction (Input View)} \\ 
& City & Valley & Forest \\
\midrule
HyFluid~\cite{yu2024inferring} & 23.24 & 12.43 & 13.70 \\
FluidNexus~\cite{gao2025fluidnexus} & 24.18 & 16.44 &  14.63 \\
Ours & \textbf{24.68} & \textbf{20.42} &  \textbf{17.91}\\ \bottomrule
\end{tabular}
}

\label{table:pixabay}
\end{table}

\subsection{Smoke Videos in the Wild}
\label{sec:exp_real}

We further evaluate on in-the-wild smoke videos.
Due to the lack of ground-truth novel views in real videos, here we evaluate only the task of \textit{future prediction at input view}, meaning the camera poses along the source video.
\vspace{-1 em}
\paragraph{FLAME~\cite{shamsoshoara2021aerial}.}
The FLAME dataset includes drone-captured videos in the wild forest.
The smoke is light, and thus background removal is necessary.
We show the results in Table~\ref{table:flame} and Figure~\ref{fig:flame_visualization}.
\vspace{-1 em}
\paragraph{\href{https://pixabay.com/}{Pixabay}.}
We evaluate three thick-smoke videos from Pixabay.
We show the results in Table~\ref{table:pixabay} and Figure~\ref{fig:pixabay_visualization}.

\vspace{1 em}
Overall, across both light and dense smoke over diverse real-world videos, our method outperforms prior work, achieving an average PSNR improvement of +2.22 dB.
Qualitatively, the reconstructions remain consistent and blend back into the original backgrounds.

\begin{figure}[h]
\centering
\includegraphics[width=.8\linewidth]{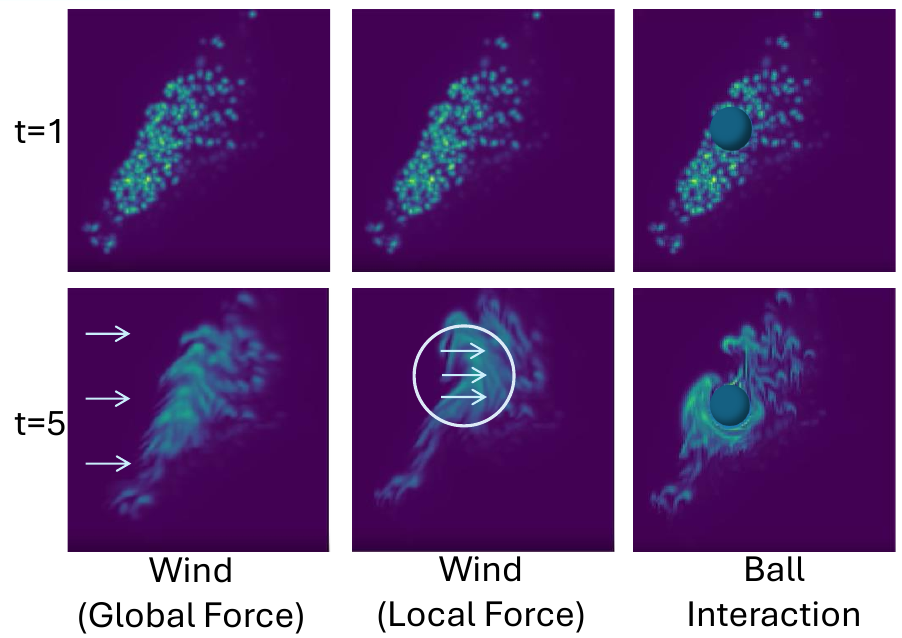}
\centering 
\vspace{-0.5em}
\captionsetup{font=small}
\caption{Visualizations of simulations with our smoke assets in different interaction scenarios. $t$ means simulation time, different from the video frames we used in previous figures. The simulation starts with particles reconstructed at the last frame of the ``city'' video from Pixabay (Figure~\ref{fig:pixabay_visualization} top).}
\vspace{-1em}
\label{fig:simulation_visualization}
\end{figure}

\begin{table*}[t!]
\centering
\captionsetup{font=small}
\vspace{-0.5 em}
\caption{Video resolutions ($W\times H \times T$) and costs of training and inference of our pipeline.}
\vspace{-0.5 em}
\resizebox{0.8\textwidth}{!}{
\addtolength{\tabcolsep}{-0.35em}
\begin{tabular}{l|cc|cc|cc}
\toprule
Dataset & \multicolumn{2}{c|}{Synthetic ($1080\times1920\times270$)} & \multicolumn{2}{c|}{FLAME ($1920\times1080\times270$)} & \multicolumn{2}{c}{Pixabay ($1920\times1080\times270$)} \\ \midrule
Stage (GPU Hours)  & Training & Inference & Training & Inference& Training & Inference\\ \midrule
Smoke Extraction & - & 0.03 & - &  0.03 & - &  0.03 \\
Background Removal & - & 0.02 & 2.06 & 0.01 & - & 0.01 \\
Multi-Views & - & 2.35 & - & 2.56 & - & 2.55 \\
Gaussian Particles & 1.03 & 0.01 & 1.17 & 0.01 & 1.76 & 0.01 \\ \midrule
Total &   \multicolumn{2}{c|}{3.44} &   \multicolumn{2}{c|}{5.84}&   \multicolumn{2}{c}{4.36}\\
\bottomrule
\end{tabular}
}
\vspace{-0.5em}
\label{table:settings}
\end{table*}

\subsection{Interactive Simulations}
\label{sec:exp_simulation}

Finally, we demonstrate interactive simulations enabled by our fluid reconstruction.
We consider two scenarios:
(i) external wind (both global and local forces),
and (ii) a rigid spherical obstacle.
For the global wind, we apply a global constant force $f_w=(0.005,0,0)$ uniformly across the domain. The local wind uses the same magnitude and direction but is confined to a sphere of radius 30 at the scene center.
For the obstacle case, we place a rigid ball (radius 10) centered at $(50, 70)$ and simulate the resulting interactions.
All simulations use the reconstruction from the ``city'' video from Pixabay (Figure~\ref{fig:pixabay_visualization} top); results are shown in Figure~\ref{fig:simulation_visualization} (XY-plane from Z-axis).
We can see that our smoke assets support realistic editing via diverse simulation scenarios.

\subsection{Training and Inference Cost}
Table \ref{table:settings} reports the per-stage GPU hours. Across datasets and resolutions, our pipeline reconstructs smoke from in-the-wild videos in 4.47 GPU-hours on average among all 5 videos.
Notably, with the DUSt3R-based pose initialization, smoke can be localized without seeding a large number of particles, reducing the Gaussian-particle training stage by 1.5 GPU-hours.

\section{Related Works}

\subsection{Fluid Field Extraction from Videos}
A series of recent works tackled fluid reconstruction from videos, but most relied on controlled, multi-view recordings that differed substantially from in-the-wild footage.
PINF~\cite{chu2022physics} coupled Navier–Stokes PDEs with a continuous spatiotemporal NeRF to recover flow;
NeuroFluid~\cite{guan2022neurofluid} introduced a particle-driven neural renderer that embedded fluid properties and a particle transition model;
HyFluid~\cite{yu2024inferring} estimated hybrid neural fields for density and velocity using physics-based losses;
FluidNexus~\cite{gao2025fluidnexus} reconstructed smoke from a single video by leveraging generative priors, yet its multi-view generator was fine-tuned on laboratory smoke with fixed, calibrated cameras and clean backgrounds.
To our knowledge, reconstructing fluid fields from a single in-the-wild video and turning them into ready-to-use dynamic 3D smoke assets remains largely underexplored.

\subsection{Physics-based Fluid Simulation and Editing}

Classical fluid simulation in computer graphics has been broadly categorized into Eulerian~\cite{souli2013arbitrary} and Lagrangian formulations~\cite{bennett2006lagrangian}. Production solvers edited flows by manipulating external forces, boundary conditions, and solid–fluid interactions (obstacles), which underpinned VFX and interactive applications. Recent differentiable-physics systems such as DiffTaichi~\cite{hu2019difftaichi} and PhiFlow~\cite{holl2024phiflow} made long-horizon, gradient-based editing practical.
These tools provided useful platforms for controlled physical interactions.
Building on these advances, our work reconstructs smoke directly from in-the-wild videos and integrates simulation for realistic editing, extending such physics-based editing to unconstrained real-world scenarios.

\subsection{4D Generation}

Recent advances in generative modeling have markedly improved image-to-3D and image-to-novel-view-synthesis, and video diffusion models with stabilized virtual cameras produced short clips with a degree of multi-view consistency~\cite{zhou2025stable,liu2023zero,yu2024viewcrafter}.
SV4D~\cite{xie2024sv4d} and SV4D~2.0~\cite{yao2025sv4d} extended view-consistent generation to dynamic scenes, reinforcing temporal continuity.
In our pipeline, we build on this generative consistency to obtain auxiliary multi-view trajectories, but go further by reconstructing physically plausible dynamic smoke fields.

\section{Conclusion}
In this paper, we addressed the challenge of recovering time-varying 3D smoke fields from a single in-the-wild video and producing ready-to-use dynamic assets. The proposed pipeline tackles three core challenges—noisy backgrounds, unknown camera poses, and coupled spatiotemporal views—through one-shot segmentation and dehazing, pose estimation, and decoupled spatiotemporal trajectory through multi-view generation and local pose perturbation. We further showed that the reconstructed assets support physically consistent editing through fluid simulation, aligning visual fidelity with physical plausibility and supporting downstream simulation workflows.

\section*{Acknowledgement}

We thank Dr. Jiajun Wu and Hong-Xing ``Koven'' Yu for helpful comments and suggestions.

{
    \small
\bibliographystyle{ieeenat_fullname}
\bibliography{main}

\begin{thebibliography}{44}
\providecommand{\natexlab}[1]{#1}
\providecommand{\url}[1]{\texttt{#1}}
\expandafter\ifx\csname urlstyle\endcsname\relax
  \providecommand{\doi}[1]{doi: #1}\else
  \providecommand{\doi}{doi: \begingroup \urlstyle{rm}\Url}\fi

\bibitem[Bai et~al.(2020)Bai, Li, Desbrun, and Liu]{bai2020dynamic}
Kai Bai, Wei Li, Mathieu Desbrun, and Xiaopei Liu.
\newblock Dynamic upsampling of smoke through dictionary-based learning.
\newblock \emph{ACM Transactions on Graphics (TOG)}, 40\penalty0 (1):\penalty0 1--19, 2020.

\bibitem[Baker et~al.(2024)Baker, Hobley, Scherl, Fang, Leach, and Davy]{baker2024enginebench}
Samuel~J Baker, Michael~A Hobley, Isabel Scherl, Xiaohang Fang, Felix~CP Leach, and Martin~H Davy.
\newblock Enginebench: flow reconstruction in the transparent combustion chamber iii optical engine.
\newblock \emph{arXiv preprint arXiv:2406.03325}, 2024.

\bibitem[Bennett(2006)]{bennett2006lagrangian}
Andrew Bennett.
\newblock \emph{Lagrangian fluid dynamics}.
\newblock Cambridge University Press, 2006.

\bibitem[Chu et~al.(2021)Chu, Thuerey, Seidel, Theobalt, and Zayer]{chu2021learning}
Mengyu Chu, Nils Thuerey, Hans-Peter Seidel, Christian Theobalt, and Rhaleb Zayer.
\newblock Learning meaningful controls for fluids.
\newblock \emph{ACM Transactions on Graphics (TOG)}, 40\penalty0 (4):\penalty0 1--13, 2021.

\bibitem[Chu et~al.(2022)Chu, Liu, Zheng, Franz, Seidel, Theobalt, and Zayer]{chu2022physics}
Mengyu Chu, Lingjie Liu, Quan Zheng, Erik Franz, Hans-Peter Seidel, Christian Theobalt, and Rhaleb Zayer.
\newblock Physics informed neural fields for smoke reconstruction with sparse data.
\newblock \emph{ACM Transactions on Graphics (ToG)}, 41\penalty0 (4):\penalty0 1--14, 2022.

\bibitem[Deng et~al.(2023{\natexlab{a}})Deng, Yu, Wu, and Zhu]{deng2023learning}
Yitong Deng, Hong-Xing Yu, Jiajun Wu, and Bo Zhu.
\newblock Learning vortex dynamics for fluid inference and prediction.
\newblock \emph{arXiv preprint arXiv:2301.11494}, 2023{\natexlab{a}}.

\bibitem[Deng et~al.(2023{\natexlab{b}})Deng, Yu, Zhang, Wu, and Zhu]{deng2023fluid}
Yitong Deng, Hong-Xing Yu, Diyang Zhang, Jiajun Wu, and Bo Zhu.
\newblock Fluid simulation on neural flow maps.
\newblock \emph{ACM Transactions on Graphics (TOG)}, 42\penalty0 (6):\penalty0 1--21, 2023{\natexlab{b}}.

\bibitem[Eckert et~al.(2019)Eckert, Um, and Thuerey]{eckert2019scalarflow}
Marie-Lena Eckert, Kiwon Um, and Nils Thuerey.
\newblock Scalarflow: a large-scale volumetric data set of real-world scalar transport flows for computer animation and machine learning.
\newblock \emph{ACM Transactions on Graphics (TOG)}, 38\penalty0 (6):\penalty0 1--16, 2019.

\bibitem[et~al.(2018)]{kupyn2018deblurgan}
Kupyn et al.
\newblock Deblurgan: Blind motion deblurring using conditional adversarial networks.
\newblock In \emph{CVPR}, 2018.

\bibitem[et~al.(2017)]{nah2017deep}
Nah et al.
\newblock Deep multi-scale convolutional neural network for dynamic scene deblurring.
\newblock In \emph{CVPR}, 2017.

\bibitem[Gao et~al.(2025)Gao, Yu, Zhu, and Wu]{gao2025fluidnexus}
Yue Gao, Hong-Xing Yu, Bo Zhu, and Jiajun Wu.
\newblock Fluidnexus: 3d fluid reconstruction and prediction from a single video.
\newblock In \emph{Proceedings of the Computer Vision and Pattern Recognition Conference}, pages 26091--26101, 2025.

\bibitem[Guan et~al.(2022)Guan, Deng, Wang, and Yang]{guan2022neurofluid}
Shanyan Guan, Huayu Deng, Yunbo Wang, and Xiaokang Yang.
\newblock Neurofluid: Fluid dynamics grounding with particle-driven neural radiance fields.
\newblock In \emph{International Conference on Machine Learning}, pages 7919--7929. PMLR, 2022.

\bibitem[He et~al.(2010)He, Sun, and Tang]{he2010single}
Kaiming He, Jian Sun, and Xiaoou Tang.
\newblock Single image haze removal using dark channel prior.
\newblock \emph{IEEE transactions on pattern analysis and machine intelligence}, 33\penalty0 (12):\penalty0 2341--2353, 2010.

\bibitem[Holl and Thuerey(2024)]{holl2024phiflow}
Philipp Holl and Nils Thuerey.
\newblock ${\Phi}_{\text{flow}}$ ({PhiFlow}): Differentiable simulations for pytorch, tensorflow and jax.
\newblock In \emph{International Conference on Machine Learning}. PMLR, 2024.

\bibitem[Hu et~al.(2019)Hu, Li, Anderson, Ragan-Kelley, and Durand]{hu2019taichi}
Yuanming Hu, Tzu-Mao Li, Luke Anderson, Jonathan Ragan-Kelley, and Fr{\'e}do Durand.
\newblock Taichi: a language for high-performance computation on spatially sparse data structures.
\newblock \emph{ACM Transactions on Graphics (TOG)}, 38\penalty0 (6):\penalty0 201, 2019.

\bibitem[Hu et~al.(2020)Hu, Anderson, Li, Sun, Carr, Ragan-Kelley, and Durand]{hu2019difftaichi}
Yuanming Hu, Luke Anderson, Tzu-Mao Li, Qi Sun, Nathan Carr, Jonathan Ragan-Kelley, and Fr{\'e}do Durand.
\newblock Difftaichi: Differentiable programming for physical simulation.
\newblock \emph{ICLR}, 2020.

\bibitem[Hu et~al.(2021)Hu, Liu, Yang, Xu, Kuang, Xu, Dai, Freeman, and Durand]{hu2021quantaichi}
Yuanming Hu, Jiafeng Liu, Xuanda Yang, Mingkuan Xu, Ye Kuang, Weiwei Xu, Qiang Dai, William~T. Freeman, and Frédo Durand.
\newblock Quantaichi: A compiler for quantized simulations.
\newblock \emph{ACM Transactions on Graphics (TOG)}, 40\penalty0 (4), 2021.

\bibitem[Kerbl et~al.(2023)Kerbl, Kopanas, Leimk{\"u}hler, and Drettakis]{kerbl20233d}
Bernhard Kerbl, Georgios Kopanas, Thomas Leimk{\"u}hler, and George Drettakis.
\newblock 3d gaussian splatting for real-time radiance field rendering.
\newblock \emph{ACM Trans. Graph.}, 42\penalty0 (4):\penalty0 139--1, 2023.

\bibitem[Kim et~al.(2020)Kim, Azevedo, Gross, and Solenthaler]{kim2020lagrangian}
Byungsoo Kim, Vinicius~C Azevedo, Markus Gross, and Barbara Solenthaler.
\newblock Lagrangian neural style transfer for fluids.
\newblock \emph{ACM Transactions on Graphics (TOG)}, 39\penalty0 (4):\penalty0 52--1, 2020.

\bibitem[Kirillov et~al.(2023)Kirillov, Mintun, Ravi, Mao, Rolland, Gustafson, Xiao, Whitehead, Berg, Lo, et~al.]{kirillov2023segment}
Alexander Kirillov, Eric Mintun, Nikhila Ravi, Hanzi Mao, Chloe Rolland, Laura Gustafson, Tete Xiao, Spencer Whitehead, Alexander~C Berg, Wan-Yen Lo, et~al.
\newblock Segment anything.
\newblock In \emph{Proceedings of the IEEE/CVF international conference on computer vision}, pages 4015--4026, 2023.

\bibitem[Liu et~al.(2023)Liu, Wu, Van~Hoorick, Tokmakov, Zakharov, and Vondrick]{liu2023zero}
Ruoshi Liu, Rundi Wu, Basile Van~Hoorick, Pavel Tokmakov, Sergey Zakharov, and Carl Vondrick.
\newblock Zero-1-to-3: Zero-shot one image to 3d object.
\newblock In \emph{Proceedings of the IEEE/CVF international conference on computer vision}, pages 9298--9309, 2023.

\bibitem[Macklin and M{\"u}ller(2013)]{macklin2013position}
Miles Macklin and Matthias M{\"u}ller.
\newblock Position based fluids.
\newblock \emph{ACM Transactions on Graphics (TOG)}, 32\penalty0 (4):\penalty0 1--12, 2013.

\bibitem[Macklin et~al.(2014)Macklin, M{\"u}ller, Chentanez, and Kim]{macklin2014unified}
Miles Macklin, Matthias M{\"u}ller, Nuttapong Chentanez, and Tae-Yong Kim.
\newblock Unified particle physics for real-time applications.
\newblock \emph{ACM Transactions on Graphics (TOG)}, 33\penalty0 (4):\penalty0 1--12, 2014.

\bibitem[M{\"u}ller et~al.(2003)M{\"u}ller, Charypar, and Gross]{muller2003particle}
Matthias M{\"u}ller, David Charypar, and Markus Gross.
\newblock Particle-based fluid simulation for interactive applications.
\newblock In \emph{Proceedings of the 2003 ACM SIGGRAPH/Eurographics symposium on Computer animation}, pages 154--159, 2003.

\bibitem[Poole et~al.(2022)Poole, Jain, Barron, and Mildenhall]{poole2022dreamfusion}
Ben Poole, Ajay Jain, Jonathan~T Barron, and Ben Mildenhall.
\newblock Dreamfusion: Text-to-3d using 2d diffusion.
\newblock \emph{arXiv preprint arXiv:2209.14988}, 2022.

\bibitem[Saini et~al.(2016)Saini, Arndt, and Steinberg]{saini2016development}
Pankaj Saini, Christoph~M Arndt, and Adam~M Steinberg.
\newblock Development and evaluation of gappy-pod as a data reconstruction technique for noisy piv measurements in gas turbine combustors.
\newblock \emph{Experiments in Fluids}, 57\penalty0 (7):\penalty0 122, 2016.

\bibitem[Selle et~al.(2008)Selle, Fedkiw, Kim, Liu, and Rossignac]{selle2008unconditionally}
Andrew Selle, Ronald Fedkiw, Byungmoon Kim, Yingjie Liu, and Jarek Rossignac.
\newblock An unconditionally stable maccormack method.
\newblock \emph{Journal of Scientific Computing}, 35\penalty0 (2):\penalty0 350--371, 2008.

\bibitem[Shamsoshoara et~al.(2021)Shamsoshoara, Afghah, Razi, Zheng, Ful{\'e}, and Blasch]{shamsoshoara2021aerial}
Alireza Shamsoshoara, Fatemeh Afghah, Abolfazl Razi, Liming Zheng, Peter~Z Ful{\'e}, and Erik Blasch.
\newblock Aerial imagery pile burn detection using deep learning: The flame dataset.
\newblock \emph{Computer Networks}, 193:\penalty0 108001, 2021.

\bibitem[Song et~al.(2023)Song, He, Qian, and Du]{song2023vision}
Yuda Song, Zhuqing He, Hui Qian, and Xin Du.
\newblock Vision transformers for single image dehazing.
\newblock \emph{IEEE Transactions on Image Processing}, 32:\penalty0 1927--1941, 2023.

\bibitem[Souli and Benson(2013)]{souli2013arbitrary}
M'hamed Souli and David~J Benson.
\newblock \emph{Arbitrary Lagrangian Eulerian and fluid-structure interaction: numerical simulation}.
\newblock John Wiley \& Sons, 2013.

\bibitem[Thuerey et~al.(2020)Thuerey, Wei{\ss}enow, Prantl, and Hu]{thuerey2020deep}
Nils Thuerey, Konstantin Wei{\ss}enow, Lukas Prantl, and Xiangyu Hu.
\newblock Deep learning methods for reynolds-averaged navier--stokes simulations of airfoil flows.
\newblock \emph{AIAA Journal}, 58\penalty0 (1):\penalty0 25--36, 2020.

\bibitem[Wang et~al.(2024{\natexlab{a}})Wang, Leroy, Cabon, Chidlovskii, and Revaud]{wang2024dust3r}
Shuzhe Wang, Vincent Leroy, Yohann Cabon, Boris Chidlovskii, and Jerome Revaud.
\newblock Dust3r: Geometric 3d vision made easy.
\newblock In \emph{Proceedings of the IEEE/CVF Conference on Computer Vision and Pattern Recognition}, pages 20697--20709, 2024{\natexlab{a}}.

\bibitem[Wang et~al.(2023)Wang, Wang, Cao, Shen, and Huang]{wang2023images}
Xinlong Wang, Wen Wang, Yue Cao, Chunhua Shen, and Tiejun Huang.
\newblock Images speak in images: A generalist painter for in-context visual learning.
\newblock In \emph{Proceedings of the IEEE/CVF Conference on Computer Vision and Pattern Recognition}, pages 6830--6839, 2023.

\bibitem[Wang et~al.(2024{\natexlab{b}})Wang, Xu, Liu, Ren, Kosinka, Telea, Wang, Song, Chang, Li, et~al.]{wang2024physics}
Xiaokun Wang, Yanrui Xu, Sinuo Liu, Bo Ren, Jiri Kosinka, Alexandru~C Telea, Jiamin Wang, Chongming Song, Jian Chang, Chenfeng Li, et~al.
\newblock Physics-based fluid simulation in computer graphics: Survey, research trends, and challenges.
\newblock \emph{Computational Visual Media}, pages 1--56, 2024{\natexlab{b}}.

\bibitem[Xiang et~al.(2024)Xiang, Lv, Xu, Deng, Wang, Zhang, Chen, Tong, and Yang]{xiang2024structured}
Jianfeng Xiang, Zelong Lv, Sicheng Xu, Yu Deng, Ruicheng Wang, Bowen Zhang, Dong Chen, Xin Tong, and Jiaolong Yang.
\newblock Structured 3d latents for scalable and versatile 3d generation.
\newblock \emph{arXiv preprint arXiv:2412.01506}, 2024.

\bibitem[Xie et~al.(2024)Xie, Yao, Voleti, Jiang, and Jampani]{xie2024sv4d}
Yiming Xie, Chun-Han Yao, Vikram Voleti, Huaizu Jiang, and Varun Jampani.
\newblock Sv4d: Dynamic 3d content generation with multi-frame and multi-view consistency.
\newblock \emph{arXiv preprint arXiv:2407.17470}, 2024.

\bibitem[Yao et~al.(2025)Yao, Xie, Voleti, Jiang, and Jampani]{yao2025sv4d}
Chun-Han Yao, Yiming Xie, Vikram Voleti, Huaizu Jiang, and Varun Jampani.
\newblock Sv4d 2.0: Enhancing spatio-temporal consistency in multi-view video diffusion for high-quality 4d generation.
\newblock \emph{arXiv preprint arXiv:2503.16396}, 2025.

\bibitem[Yu et~al.(2024{\natexlab{a}})Yu, Zheng, Gao, Deng, Zhu, and Wu]{yu2024inferring}
Hong-Xing Yu, Yang Zheng, Yuan Gao, Yitong Deng, Bo Zhu, and Jiajun Wu.
\newblock Inferring hybrid neural fluid fields from videos.
\newblock \emph{Advances in Neural Information Processing Systems}, 36, 2024{\natexlab{a}}.

\bibitem[Yu et~al.(2024{\natexlab{b}})Yu, Xing, Yuan, Hu, Li, Huang, Gao, Wong, Shan, and Tian]{yu2024viewcrafter}
Wangbo Yu, Jinbo Xing, Li Yuan, Wenbo Hu, Xiaoyu Li, Zhipeng Huang, Xiangjun Gao, Tien-Tsin Wong, Ying Shan, and Yonghong Tian.
\newblock Viewcrafter: Taming video diffusion models for high-fidelity novel view synthesis.
\newblock \emph{arXiv preprint arXiv:2409.02048}, 2024{\natexlab{b}}.

\bibitem[Zang et~al.(2020)Zang, Idoughi, Wang, Bennett, Du, Skeen, Roberts, Wonka, and Heidrich]{zang2020tomofluid}
Guangming Zang, Ramzi Idoughi, Congli Wang, Anthony Bennett, Jianguo Du, Scott Skeen, William~L Roberts, Peter Wonka, and Wolfgang Heidrich.
\newblock Tomofluid: Reconstructing dynamic fluid from sparse view videos.
\newblock In \emph{Proceedings of the IEEE/CVF Conference on Computer Vision and Pattern Recognition}, pages 1870--1879, 2020.

\bibitem[Zhang et~al.(2018)Zhang, Isola, Efros, Shechtman, and Wang]{zhang2018unreasonable}
Richard Zhang, Phillip Isola, Alexei~A Efros, Eli Shechtman, and Oliver Wang.
\newblock The unreasonable effectiveness of deep features as a perceptual metric.
\newblock In \emph{Proceedings of the IEEE conference on computer vision and pattern recognition}, pages 586--595, 2018.

\bibitem[Zhao et~al.(2025{\natexlab{a}})Zhao, Lai, Lin, Zhao, Liu, Yang, Feng, Yang, Zhang, Yang, et~al.]{zhao2025hunyuan3d}
Zibo Zhao, Zeqiang Lai, Qingxiang Lin, Yunfei Zhao, Haolin Liu, Shuhui Yang, Yifei Feng, Mingxin Yang, Sheng Zhang, Xianghui Yang, et~al.
\newblock Hunyuan3d 2.0: Scaling diffusion models for high resolution textured 3d assets generation.
\newblock \emph{arXiv preprint arXiv:2501.12202}, 2025{\natexlab{a}}.

\bibitem[Zhao et~al.(2025{\natexlab{b}})Zhao, Zhao, Li, and Hu]{zhao2025vid2fluid}
Zhiwei Zhao, Alan Zhao, Minchen Li, and Yixin Hu.
\newblock Vid2fluid: 3d dynamic fluid assets from single-view videos with generative gaussian splatting.
\newblock \emph{arXiv preprint arXiv:2503.00868}, 2025{\natexlab{b}}.

\bibitem[Zhou et~al.(2025)Zhou, Gao, Voleti, Vasishta, Yao, Boss, Torr, Rupprecht, and Jampani]{zhou2025stable}
Jensen Zhou, Hang Gao, Vikram Voleti, Aaryaman Vasishta, Chun-Han Yao, Mark Boss, Philip Torr, Christian Rupprecht, and Varun Jampani.
\newblock Stable virtual camera: Generative view synthesis with diffusion models.
\newblock \emph{arXiv preprint arXiv:2503.14489}, 2025.

\end{thebibliography}
}


\appendix
\clearpage
\setcounter{page}{1}
\maketitlesupplementary

\section{Technical Details}
\label{sec:tech_supp}
\subsection{Synthetic Data} 
\label{sec:syn_data_supp}
\begin{figure}[h]
\centering
\includegraphics[width=1.0\linewidth]{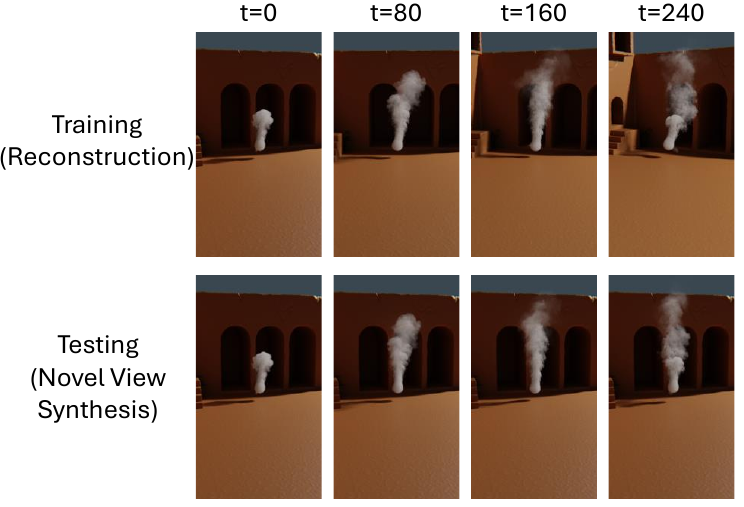}
\centering 
\vspace{-0.5em}
\captionsetup{font=small}
\caption{Frames in our synthetic smoke video.
When training the reconstruction, the camera view is moving along a trajectory defined by Equation~\ref{eq:synthetic_camera_trajectory}.
During testing, we evaluate the novel view synthesis with the fixed camera pose from the first frame ($\phi(t=0)$).
} 
\vspace{-1em}
\label{fig:synthetic_samples}
\end{figure}

To construct the synthetic dataset for our evaluation on novel view synthesis, we obtain a 4D smoke sequence and a static 3D scene from \href{https://www.cgtrader.com/3d-models}{CGTrader}.
The sequence is placed into the 3D scene and rendered along a prescribed camera path.
For the training video, the camera pitch is fixed at \(\theta_p = 10^\circ\),
and the azimuth is swept linearly from \(-105^\circ\) to \(-45^\circ\) at a constant speed:
\begin{equation}
\phi(t) \,=\, -105^\circ \;+\; 60^\circ \cdot \frac{t}{T'}, \quad t\in[0,\,T'],
\label{eq:synthetic_camera_trajectory}
\end{equation}
where $T'=270$.
Rendering is performed with Blender \textit{Cycles} (max samples \(=200\), i.e., each pixel is estimated by averaging up to 200 light paths to reduce Monte Carlo noise).
Representative frames are shown in Figure~\ref{fig:synthetic_samples}.
In the testing video for the novel view synthesis, the camera is fixed at
\(\phi(t=0)=-105^\circ\) and \(\theta_p=10^\circ\), i.e., the camera pose at the first frame.
During inference, we use the
pose estimated by DUSt3R at the first frame from the training video as the input pose.

\subsection{Pose Estimation}
\label{sec:pose_estimation_supp}
\begin{figure}[h]
\centering
\includegraphics[width=1.0\linewidth]{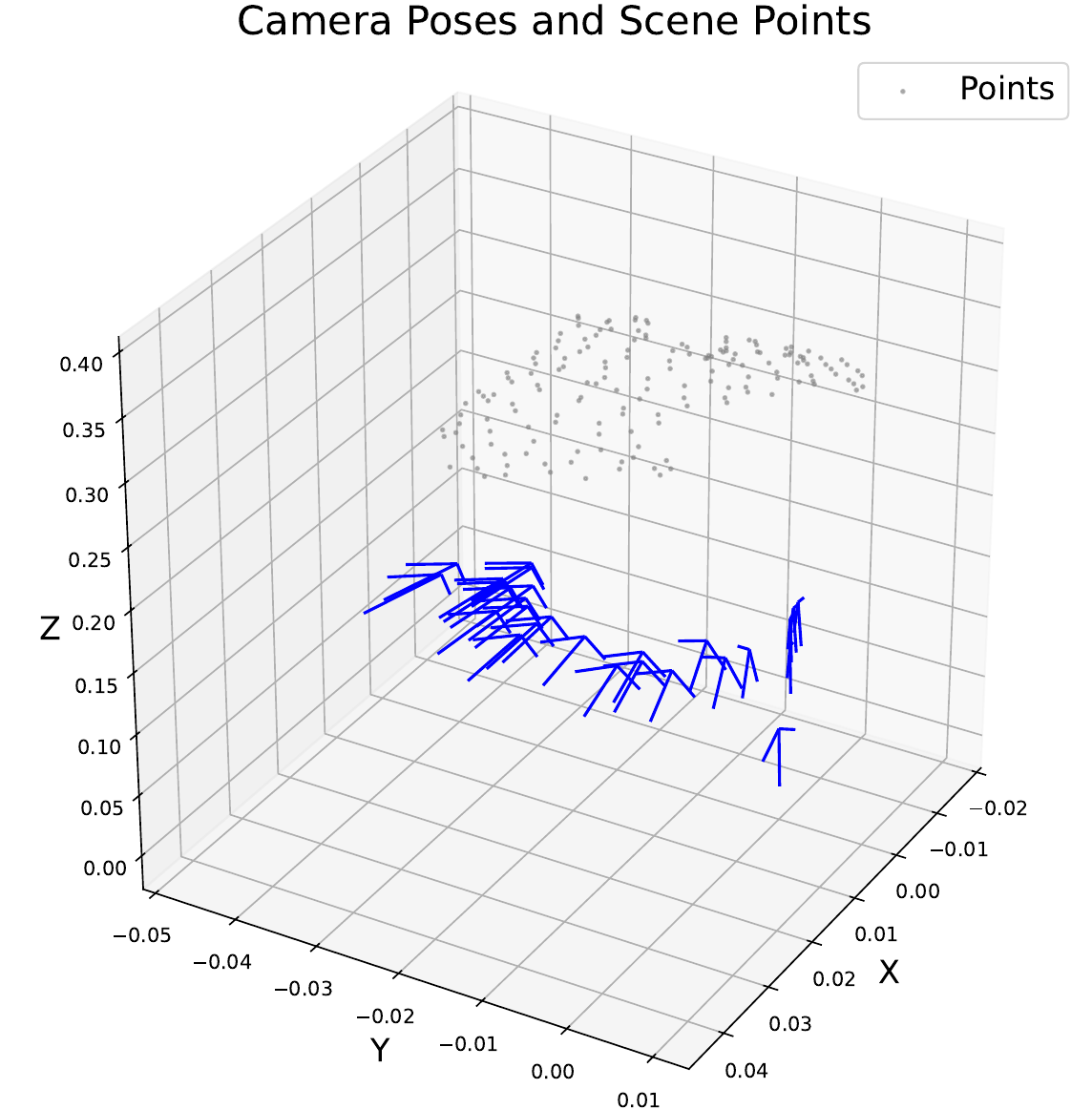}
\centering 
\vspace{-0.5em}
\captionsetup{font=small}
\caption{Camera poses and foreground smoke points (for the ``valley'' video in Figure~\ref{fig:pixabay_visualization} (middle) from Pixabay) predicted by DUSt3R. Blue arrows are DUSt3R camera-to-world poses, gray dots are the reconstructed foreground point clouds.}
\vspace{-1em}
\label{fig:poses_vis}
\end{figure}

We infer per-frame camera poses with a pretrained DUSt3R in the one-reference mode,
using the first frame as reference. Since DUSt3R and Gaussian Splatting (GS) use
different conventions (GS looks along $-Z$), we convert DUSt3R camera-to-world poses by
\emph{negating} the $y$ and $z$ axes (i.e., multiply both by $-1$). An example of estimated poses and point cloud is shown in Figure \ref{fig:poses_vis}.
Formally, for each DUSt3R-predicted pose $\mathbf{C}$, we apply
\[
\mathbf{C}' = \mathbf{F}\,\mathbf{C}, \quad 
\mathbf{F} = \mathrm{diag}(1,-1,-1,1),
\]
where $\mathbf{F}$ flips the $y$ and $z$ axes.

\subsection{Particle Initialization}
\label{sec:particle_init_supp}
To accelerate convergence, we initialize both physical and visual particles from the foreground point cloud predicted by DUSt3R.
Let \(\mathbf{X}\in\mathbb{R}^{T\times H\times W\times 3}\) denote the 3D points (camera-to-world coordinates) for all frames, where \(\mathbf{X}_{t,h,w}\in\mathbb{R}^3\) is the 3D position of pixel \((h,w)\) at time \(t\).
We obtain a reliable smoke foreground by intersecting the segmentation masks \(M_t\) with the DUSt3R confidence mask \(M^{\text{conf}}_t\), and collect the foreground set:
$$
\mathcal{P}
=\Bigl\{\mathbf{x}\in\mathbb{R}^3 \;\Big|\; \mathbf{x}=\mathbf{X}_{t,h,w}, \;
M_t(h,w)\wedge M^{\text{conf}}_t(h,w)=1 \Bigr\}.
$$
As DUSt3R and Gaussian Splatting (GS) follow different axis conventions (GS looks along \(-Z\)), we convert point coordinates by \emph{negating} the \(y\) and \(z\) axes (multiply both by \(-1\), not swapping them).
With \(\mathbf{F'}=\mathrm{diag}(1,-1,-1)\), we define
$$
\tilde{\mathcal{P}}
=\Bigl\{\mathbf{F’}\,\mathbf{x}
\;\Big|\; \forall\,\mathbf{x}\in\mathcal{P} \Bigr\}.
$$
After that, we downsample \(\tilde{\mathcal{P}}\) with a voxel grid, which merges points falling into the same 3D cell into a single representative, to control the initial particle count.
We retain \(100\text{--}300\) foreground points per video.
The coordinates of the points are then used to initialize our physical and visual particles.

\subsection{Generated Videos}
\label{sec:generated_vid_supp}
SV4D 2.0 takes yaw/pitch in radians, so we recover each frame’s camera pose by applying the generation-time angle offsets to the DUSt3R pose, while using the DUSt3R-predicted point cloud center as the fixed look-at target.
Let the DUSt3R pose of the input frame be
\(\mathbf{P}_0=[\mathbf{R}_0\,|\,\mathbf{t}_0]\in\mathbb{R}^{3\times4}\),
and let \(\mathbf{c}\in\mathbb{R}^3\) be the scene center estimated from the DUSt3R point cloud.
Given per-frame angle offsets (in radians) \(\Delta\phi_t\) (yaw/azimuth) and \(\Delta\theta_t\) (pitch), define
\[
\mathbf{R}_y(\Delta\phi)=
\begin{bmatrix}
\cos\Delta\phi & 0 & \sin\Delta\phi\\
0 & 1 & 0\\
-\sin\Delta\phi & 0 & \cos\Delta\phi
\end{bmatrix},\]
\[
\mathbf{R}_x(\Delta\theta)=
\begin{bmatrix}
1 & 0 & 0\\
0 & \cos\Delta\theta & -\sin\Delta\theta\\
0 & \sin\Delta\theta & \cos\Delta\theta
\end{bmatrix}.
\]
We apply the world-space rotation \(\Delta\mathbf{R}_t=\mathbf{R}_y(\Delta\phi_t)\,\mathbf{R}_x(\Delta\theta_t)\) about pivot \(\mathbf{c}\).
The pose of frame \(t\) is:
$$\mathbf{R}_t=\Delta\mathbf{R}_t\,\mathbf{R}_0,  \mathbf{t}_t=\Delta\mathbf{R}_t\,(\mathbf{t}_0-\mathbf{c})+\mathbf{c},$$
$$\mathbf{P}_t=[\mathbf{R}_t\,|\,\mathbf{t}_t].$$

\subsection{Down-Weighting of Generated Frames}
\label{sec:weight_decay_supp}
To reduce the impact of the unreliability over time (see Section~\ref{sec:sv4d_samples}) of generated frames on reconstruction, the loss of generated frames is multiplied by an exponential decay when training visual particles:
\[
w_t \;=\; w_{\min} \;+\; (1-w_{\min})\,\exp\bigl(-k\,(t-t_0)\bigr),
\]
where \(t\) is the frame index, \(t_0=0\), \(k=0.02\), \(w_{\min}=0.0\).

\subsection{Learnable Buoyancy Strength}
In well-controlled scenes~\cite{gao2025fluidnexus}, the relative strength between buoyancy and gravity is fixed, whereas in the wild, buoyancy strength is underdetermined. To account for this, we make the buoyancy coefficient in the PBF simulation (Appendix B in~\cite{gao2025fluidnexus}) learnable and optimize it jointly with reconstruction.

\subsection{High-Frequency Information Loss} 
To sharpen fine details, we add a frequency-domain loss during visual-particle training on original input frames only (generated views are excluded due to unreliable high-frequency content).
Given an RGB frame $I$ and ground truth $\hat I$, we compute per-channel 2D FFTs
$F=\mathcal{F}\{I\}$, $\hat F=\mathcal{F}\{\hat I\}$,
and define amplitude $A=|F|$, $\hat A=|\hat F|$ and phase
$\beta=\angle F$, $\hat\beta=\angle \hat F$.
The loss is computed as the average absolute difference of amplitude and phase over all frequency bins and RGB channels:
\[
\mathcal{L}_{\mathrm{freq}}
= \operatorname{mean}\bigl(|A-\hat A|\bigr)
+ \operatorname{mean}\bigl(|\beta-\hat\beta|\bigr).
\]
We apply a linear warm-up weight
$w_t=\min\bigl(1, iter/t)$
and scale by $\lambda_{\mathrm{freq}}$:
\[
\mathcal{L}_{\mathrm{FFT}}=\lambda_{\mathrm{freq}}\; w_t\; \mathcal{L}_{\mathrm{freq}},
\]
here we set $\lambda_{\mathrm{freq}}=0.001$.

\begin{figure}[b!]
  \centering
  \includegraphics[width=\linewidth]{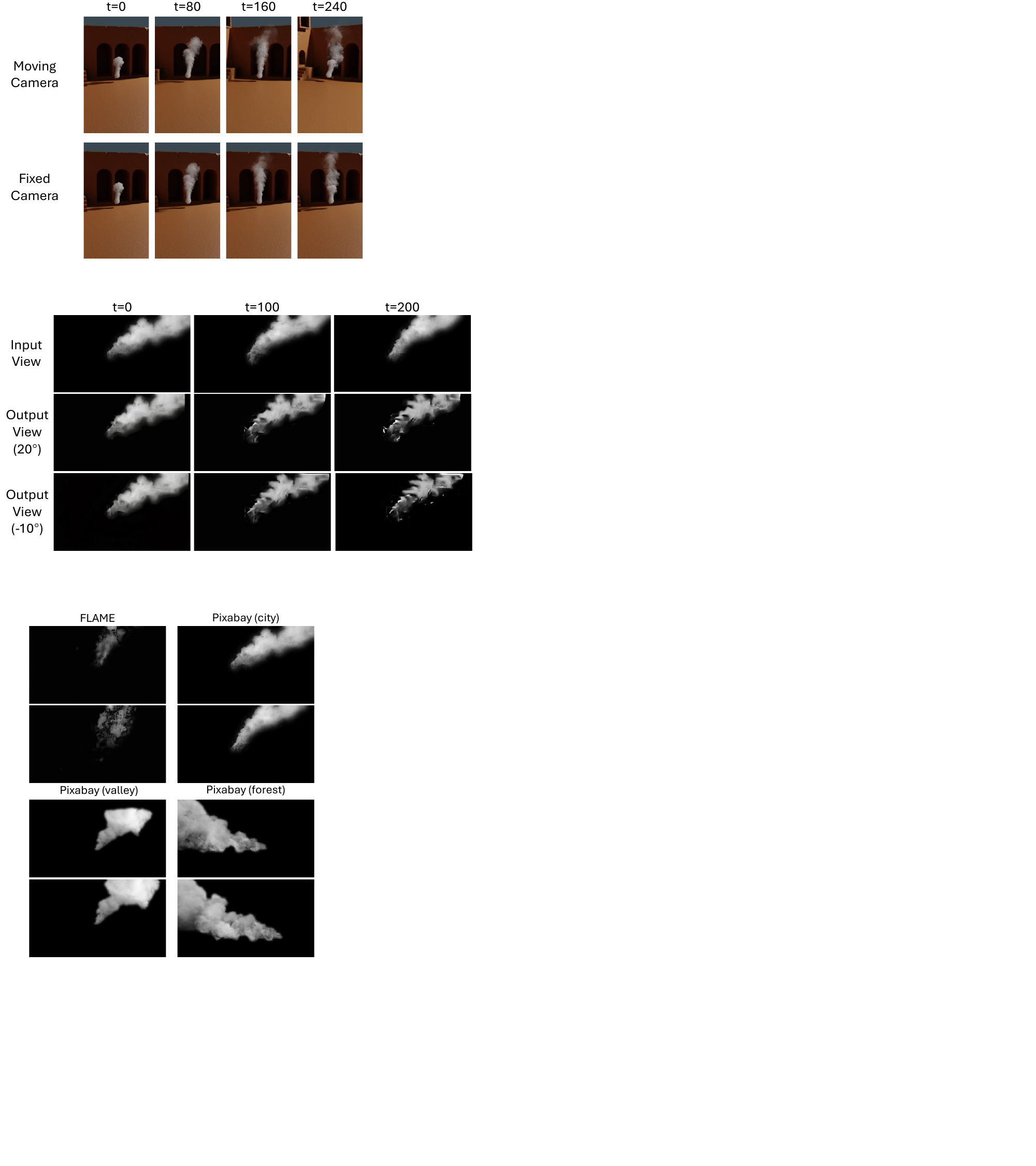}
  \caption{Visualizations of extracted smoke from wild videos.}
  \label{fig:smoke_examples}
\end{figure}

\section{More Results}
\label{sec:more_results}

\subsection{Smoke Extraction}
\label{sec:smoke_sample_supp}
\paragraph{Extracted Smoke Examples.}
Figure~\ref{fig:smoke_examples} shows representative results of our smoke extraction across several in-the-wild videos. 
For light smoke, we apply dehazing to remove background leakage; for dense smoke, we use the masked result directly as the clean foreground.

\vspace{-1.2 em}
\paragraph{Results without Smoke Extraction.}
We further train HyFluid~\cite{yu2024inferring} and FluidNexus~\cite{gao2025fluidnexus} on inputs without our smoke extraction method. The results are shown in Table~\ref{tab:no_extract_psnr}.

\begin{table}[h!]
\centering
\captionsetup{font=small}
\caption{Comparing PSNR (higher is better) of methods with/without our smoke extraction (SE) method.}
\vspace{-0.5em}
\resizebox{0.45\textwidth}{!}{
\begin{tabular}{lcccc}
\toprule
 &\multicolumn{4}{c}{Future Prediction (Input View)} \\ 
& FLAME & City & Valley & Forest \\
\midrule
HyFluid &  10.37 & 12.37 & 11.79 & 9.88  \\ 
HyFluid (w. our SE) &  21.67  & 23.24 & 12.43 & 13.70\\
FluidNexus & 6.67 & 8.78 & 9.63 & 4.46 \\ 
FluidNexus (w. our SE) & 21.78 & 24.18 & 16.44 &  14.63\\
Ours & \textbf{22.88} & \textbf{24.68} & \textbf{20.42} &  \textbf{17.91} \\ \bottomrule
\end{tabular}
}
\vspace{-0.5 em}
\label{tab:no_extract_psnr}
\end{table}

\subsection{Generated Videos from SV4D 2.0}
\label{sec:sv4d_samples}
\begin{figure}[h!]
  \centering
  \includegraphics[width=\linewidth]{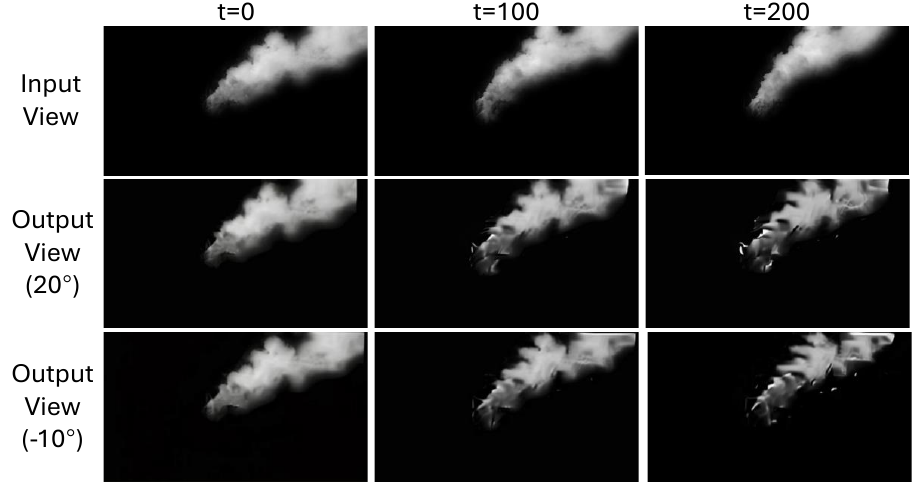}
  \caption{SV4D~2.0 multi-view generation over time. Later frames exhibit structural collapse.}
  \label{fig:sv4d2_decay}
\end{figure}

We previously noted that the quality of generated multi-view frames from SV4D~2.0 decays over time.
Figure~\ref{fig:sv4d2_decay} shows early and late frames of the sequence. As time progresses, the smoke structure gradually collapses.

\subsection{SSIM and LPIPS}
\label{sec:ssim_lpips_supp}
Beyond PSNR, we further report the structural similarity index measure (SSIM) in Table \ref{tab:ssim} and the perceptual metric LPIPS~\cite{zhang2018unreasonable} in Table \ref{tab:lpips}. Our method is still shown to perform better based on these two additional metrics.

\begin{table}[h!]
\centering
\captionsetup{font=small}
\caption{Comparing SSIM (higher is better) of smoke reconstruction by different methods on videos collected from FLAME~\cite{shamsoshoara2021aerial} and \href{https://pixabay.com/}{\ul{Pixabay}}.
}
\vspace{-0.5em}
\resizebox{0.38\textwidth}{!}{
\begin{tabular}{lcccc}
\toprule
 &\multicolumn{4}{c}{Future Prediction (Input View)} \\ 
& FLAME & City & Valley & Forest \\
\midrule
HyFluid~\cite{yu2024inferring} & 0.87 & 0.55 & 0.79 & 0.46 \\
FluidNexus~\cite{gao2025fluidnexus} & 0.88 & 0.84  & 0.76 & 0.70 \\
Ours & \textbf{0.92} & \textbf{0.89}  & \textbf{0.86} &  \textbf{0.80} \\ \bottomrule
\end{tabular}
}
\label{tab:ssim}
\end{table}

\begin{table}[h!]
\centering
\captionsetup{font=small}
\caption{Comparing LPIPS (smaller is better) of smoke reconstruction by different methods on videos collected from FLAME~\cite{shamsoshoara2021aerial} and \href{https://pixabay.com/}{\ul{Pixabay}}.
}

\vspace{-0.5em}
\resizebox{0.38\textwidth}{!}{
\begin{tabular}{lcccc}
\toprule
 &\multicolumn{4}{c}{Future Prediction (Input View)} \\ 
& FLAME & City & Valley & Forest \\
\midrule
HyFluid~\cite{yu2024inferring} &  0.13 & 0.33 &  0.18 & 0.36 \\
FluidNexus~\cite{gao2025fluidnexus} & 0.16 & 0.14  & 0.20 & 0.31 \\
Ours & \textbf{0.09} & \textbf{0.12}  & \textbf{0.13} & \textbf{0.24} \\ \bottomrule
\end{tabular}
}

\label{tab:lpips}
\end{table}

\subsection{Visualizations of Novel View Synthesis on Wild Videos}

\begin{figure}[h!]
  \centering
  \includegraphics[width=\linewidth]{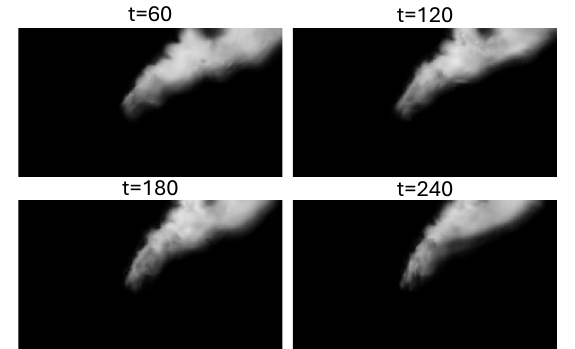}
  \caption{Novel view synthesis on wild videos.
  Our reconstructions preserve stable smoke structure across views.}
  \label{fig:nvs_wild}
\end{figure}

Figure~\ref{fig:nvs_wild} renders novel views for several wild videos. For all frames, the novel-view camera is fixed to the pose of the first frame of the training video. The reconstructed smoke remains stable over time.

\subsection{Visualizations for Ablation Study on Synthetic Data}
\label{sec:ablation_supp}
\begin{figure*}[h!]
\centering
\includegraphics[width=0.7\linewidth]{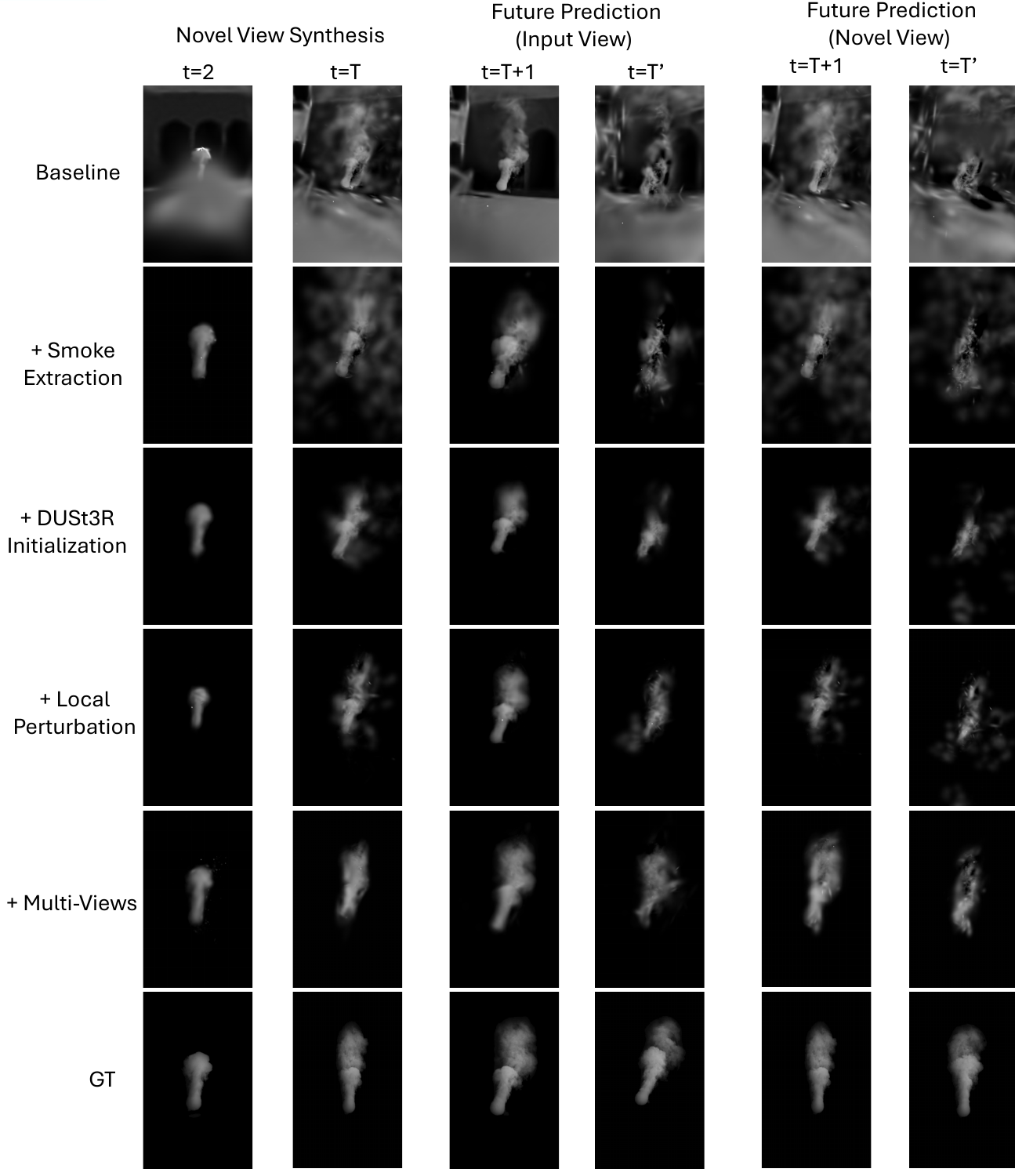}
\centering 
\vspace{-0.5em}
\captionsetup{font=small}
\caption{Ablation study of novel view synthesis and future predictions on synthetic smoke videos.
\textit{Novel view} uses the camera pose at $t=1$, and \textit{input view} means the camera poses along the source video.
Training uses $T=240$ frames, and the unseen future extends to $T'=270$ frames.
\textit{GT}: Ground Truth. } 
\vspace{-1em}
\label{fig:ablation_visualization}
\end{figure*}

We provide additional qualitative ablations (Figure~\ref{fig:ablation_visualization}) complementing the quantitative table in the main paper.

\end{document}